\documentclass[preprint,12pt]{elsarticle}

\usepackage{geometry}

\usepackage{amssymb}
\usepackage{lipsum}
\usepackage{graphicx}
\usepackage{amsmath}
\usepackage{booktabs}
\usepackage{geometry}
\usepackage{tabularx}
\usepackage{url}
\usepackage{hyperref}
\usepackage{colortbl}
\usepackage{caption}
\usepackage{subcaption}
\makeatletter
\renewcommand{\fnum@figure}{Fig. \thefigure}
\makeatother
\biboptions{sort&compress}

\begin{document}

\begin{frontmatter}


\title{MechProNet: Machine Learning Prediction of Mechanical Properties in Metal Additive Manufacturing}


\author[inst1]{Parand Akbari\footnote{Current address: Tepper School of Business, Carnegie Mellon University, Pittsburgh 15213, PA, USA }}
\affiliation[inst1]{organization={Department of Mechanical Engineering, Carnegie Mellon University},
            city={Pittsburgh},
            postcode={15213}, 
            state={PA},
            country={USA}}

\author[inst2]{Masoud Zamani}
\affiliation[inst2]{organization={Department of Mechanical Engineering and Material Science, University of Pittsburgh},
            city={Pittsburgh},
            postcode={15261}, 
            state={PA},
            country={USA}}
\author[inst3]{Amir Mostafaei}
\affiliation[inst3]{organization={Department of Mechanical, Materials, and Aerospace Engineering, Illinois Institute of Technology, 10 W 32nd Street, USA},
            city={Chicago},
            postcode={60616}, 
            state={IL},
            country={USA}}

\begin{abstract}

Predicting mechanical properties in metal additive manufacturing (MAM) is essential for ensuring the performance and reliability of printed parts, as well as their suitability for specific applications. However, conducting experiments to estimate mechanical properties in MAM processes can be laborious and expensive, and they are often limited to specific materials and processes. Machine learning (ML) methods offer a more flexible and cost-effective approach to predicting mechanical properties based on processing parameters and material properties. In this study, we introduce a comprehensive framework for benchmarking ML models for predicting mechanical properties. We compiled an extensive experimental dataset from over 90 MAM articles and data sheets from a diverse range of sources, encompassing 140 different MAM data sheets. This dataset includes information on MAM processing conditions, machines, materials, and resulting mechanical properties such as yield strength, ultimate tensile strength, elastic modulus, elongation, hardness, and surface roughness. Our framework incorporates physics-aware featurization specific to MAM, adjustable ML models, and tailored evaluation metrics to construct a comprehensive learning framework for predicting mechanical properties. Additionally, we explore the Explainable AI method, specifically SHAP analysis, to elucidate and interpret the predicted values of ML models for mechanical properties. Furthermore, data-driven explicit models were developed to estimate mechanical properties based on processing parameters and material properties, offering enhanced interpretability compared to conventional ML models.

\end{abstract}


\end{frontmatter}



\section{Introduction}
\label{sec:sample1}

Metal additive manufacturing (MAM) is a nascent technology that constructs metal components from a computer-aided design (CAD) file, typically layer by layer, through selective consolidation of feedstock material employing a heat source.\cite{DDGU,frazer} MAM has garnered significant interest from both industry and academia due to its ability to create intricate geometries, minimize material usage and lead times, and offer design flexibility, in contrast to conventional subtractive manufacturing techniques.\cite{GARDNER20232178} Due to these remarkable characteristics, it has found extensive application in advanced industries such as aerospace, biomedical, and automotive.\cite{pasang2023additive, aliyu2023laser, zhao2023direct} Nevertheless, it is crucial to prioritize the quality and reliability of additively manufactured components. \cite{cai2023review}

\vspace{3mm}

During MAM processes, defects like keyhole and lack of fusion porosities can compromise the structural integrity of the printed parts.\cite{xie2023data, laleh2023heat} Moreover, the cooling rates during and after solidification can be impacted by processing variables like beam power, scanning speed, and preheat temperature, along with material characteristics such as thermal conductivity.\cite{sneha} Additionally, the thermal cycles experienced during MAM processes alter the resulting microstructure of the component.\cite{ojo2023post,collins2016microstructural} Hence, as printed metal parts must adhere to precise mechanical property standards tailored to diverse applications across various industries, accurately predicting mechanical properties such as yield strength, ultimate tensile strength, elastic modulus, elongation at break, hardness, and roughness becomes essential to guarantee their operational effectiveness and reliability.

\vspace{3mm}

Additionally, while defects and non-equilibrium microstructures may negatively impact the mechanical properties of additively manufactured parts \cite{brennan2021defects}, their effects can be substantially mitigated through various post-processing techniques, such as heat treatments (HT) or hot isostatic pressing (HIP).\cite{laleh2023heat} However, in most cases, the mechanical properties of additively manufactured components, whether post-processed or not, can be comparable to, or even surpass, the mechanical properties of cast or wrought materials.\cite{brandl2011mechanical}

\vspace{3mm}

Examining the mechanical characteristics of components in MAM procedures through experimentation poses significant challenges due to the labor-intensive, costly, and time-consuming nature of preparation and calibration. Moreover, experiments are tailored to specific materials and processes, typically focusing on individual variables. For example, Edwards et al. \cite{edwards2013electron} explored the fatigue properties of Ti6Al4V specimens generated via electron beam additive manufacturing (EBAM) process, while Vrancken et al. \cite{vrancken2012heat} delved into the impact of heat treatment on the mechanical attributes of Ti6Al4V parts produced through selective laser melting (SLM). Additionally, Yahdollahi et al.  \cite{YADOLLAHI2015171} detailed the mechanical and microstructural attributes of 316L stainless steel fabricated via Direct Laser Deposition. Despite experimental approaches, simulation methods encounter challenges in generating precise computational forecasts regarding the relationships between processing parameters and final part quality. Furthermore, predicting mechanical properties via simulations is laborious and necessitates the integration of multiple discrete and costly simulations. \cite{yan2018data}

\vspace{3mm}

Given the constraints in experimental datasets, exploring machine learning (ML) solutions emerges as a promising avenue to address these challenges. In comparison to experimental and simulation methods, ML models offer greater flexibility and adaptability, allowing for easy adjustments as needed. Therefore, implementing ML models, constructed from experimental data, can present a more cost-efficient solution. Furthermore, the use of ML algorithms enables the examination of the collective impact of processing parameters on mechanical properties and facilitates extrapolation beyond experimental data points. Consequently, integrating ML into MAM processes not only facilitates the prediction of mechanical properties for various processing parameters but also facilitates the identification of optimal processing parameters to ensure desired mechanical properties. Recognizing these advantages, the integration of data-driven analysis and machine learning has become standard practice in advanced manufacturing and is increasingly prevalent in additive manufacturing research. Nevertheless, the application of machine learning algorithms in MAM encounters challenges due to the limited availability of highly heterogeneous and expensive-to-acquire data, which is notably smaller than datasets available for other machine learning tasks.

\vspace{3mm}

Despite these obstacles, ML has found application in MAM processes. In our prior research \cite{AKBARI2022102817}, we introduced a comprehensive framework for assessing ML in melt pool characterization, predicting melt pool geometry, as well as identifying modes and flaw types in MAM procedures. Zhan et al. \cite{ZHAN2021105941} devised a platform for data-driven analysis, focusing on fatigue life prediction of AM stainless steel 316L. They considered processing parameters such as laser power, scan speed, hatch space, and powder layer thickness effects. Xie et al. \cite{xie} developed a data-driven framework that integrates wavelet transforms and convolutional neural networks (CNN) to forecast location-dependent mechanical properties of Inconel 718 parts produced with the laser Directed Energy Deposition (L-DED) process, achieving an $R^2$ score of 0.7. Carl et al. \cite{carl} examined the efficacy of data-driven modeling, employing Ridge regression, XGBoost, and a CNN based on VGGNet to predict the yield strength of a simulated microstructural dataset of L-DED manufactured stainless steel 316L, achieving an $R^2$ value of 0.84. Xia et al. \cite{xia} utilized various ML models for surface roughness prediction of Mild carbon steel ER70S-6 parts manufactured in the wire arc additive manufacturing (WAAM) process, achieving an $R^2$ score of 0.93516. Israt et al. \cite{israt} constructed a data-driven ML model, XGBoost, to anticipate the tensile characteristics of stainless steel 316 components fabricated using the L-DED manufacturing process. Their model successfully predicted tensile properties with a limited training data size, yielding an $R^2$ score of approximately 0.7.

\vspace{3mm}
Although the aforementioned studies have effectively leveraged machine learning to enhance applications in metal additive manufacturing, their experimental datasets, processing parameters, and materials used are limited in scope. Our benchmark, however, is founded on a more extensive dataset, facilitating optimization and ensuring part quality across a broader spectrum of processing parameters and materials. Our dataset incorporates a minimum of 230 sources of experimental data \cite{1,2,3,4,5,6,7,8,9,10,11,12,13,14,15,16,17,18,19,20,21,22,23,24,25,26,27,28,29,30,31,32,33,34,35,36,37,38,39,40,41,42,43,44,45,46,47,48,49,50,51,52,53,54,55,56,57,58,59,60,61,62,63,64,65,66,67,68,69,70,71,72,73,74,75,76,77,78,79,80,81,82,83,84,87,86,edwards2013electron,vrancken2012heat, 85}, collected from experiments conducted on individual alloys produced through specific MAM processes. With this endeavor, we consolidate these diverse data sources into a larger repository for predicting mechanical properties across a wide array of processing parameters and alloys. Presently, our dataset encompasses 1600 experimental data points, spanning various processing parameters, materials, and MAM processes (PBF and DED), with ongoing expansion anticipated.

\vspace{3mm}

With our extensive dataset in hand, our aim in this initiative, known as MechProNet, is to predict the mechanical properties of additively manufactured components by crafting a suite of machine learning methodologies specifically tailored for additive manufacturing. The assessment of how various parameters associated with the building process affect the prediction performance of the employed ML models is also conducted. Furthermore, the Explainable AI approach, specifically SHAP analysis, is utilized to elucidate the prediction results of the ML models, rendering them interpretable. Additionally, an approach for model identification grounded in data is introduced to unveil the explicit connections among processing parameters, material properties, and the resulting mechanical characteristics in MAM processes. This method is designed to be more interpretable compared to conventional machine learning models.

\begin{figure}\centering
\includegraphics[scale=0.75]{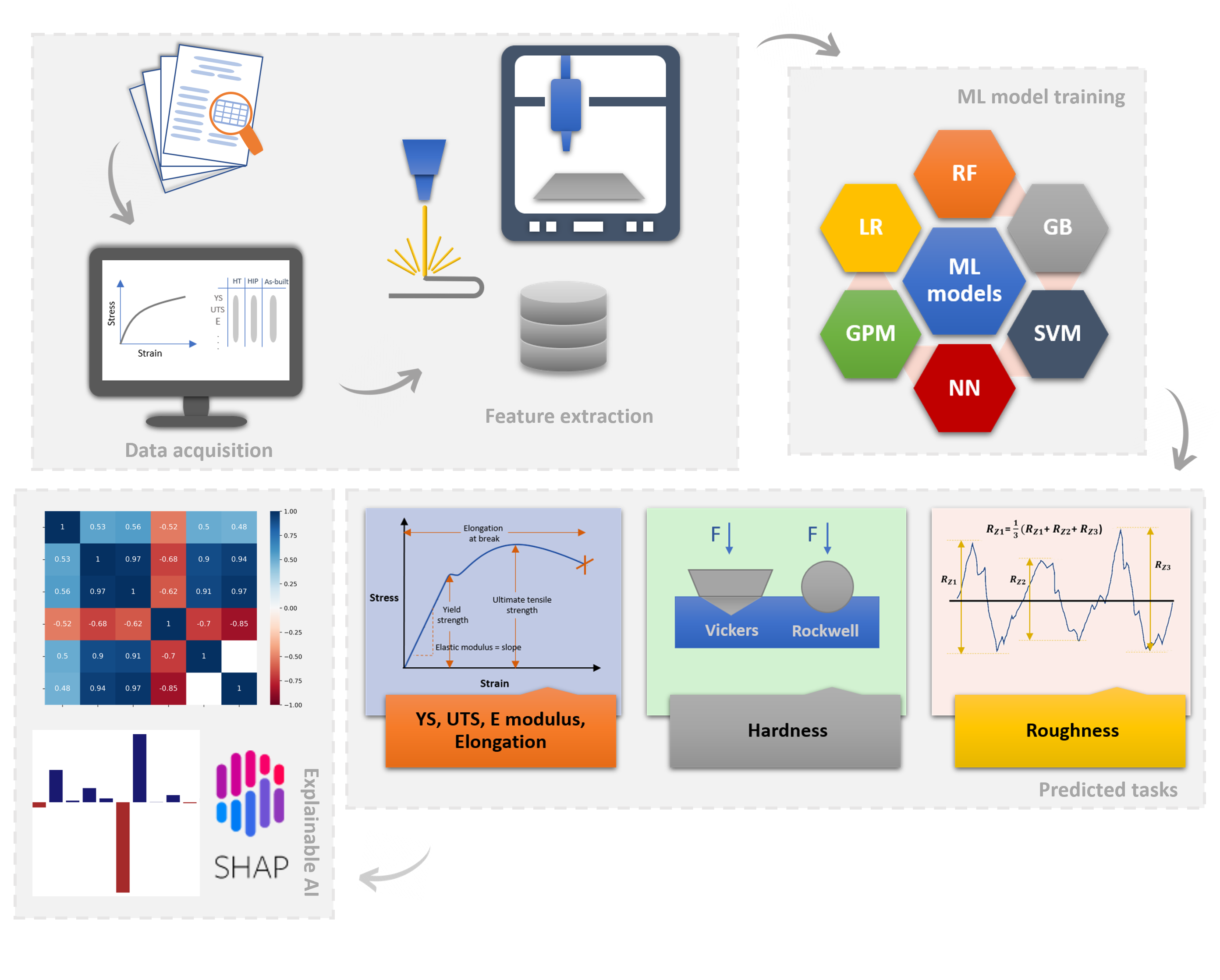}
\caption{Overview of the workflow utilized in our benchmark study, including data acquisition, feature extraction, training of machine learning models, prediction tasks, and model analysis.}
\label{fig:fig1}
\end{figure}

\section{Methodology} 

In Fig. \ref{fig:fig1}, an overview of the MechProNet framework, data collection process, dataset features, ML models used, prediction tasks, and model analysis is provided. This section will delve into the details of dataset gathering, feature engineering, and ML algorithms utilized.

\subsection{\normalfont{Data Collection}} 

We acquired empirical data from manufacturing and materials journals, as well as material datasheets from additive manufacturing companies regarding their commercial machines. Our main data sources comprised tables and figures from these publications. To ensure accuracy, the Plot Digitizer program \cite{plotdigitizer} was utilized to extract data from figures and plots. Furthermore, processing parameters and material properties were collected from each experiment to serve as input features for the ML models used in our study.

\subsection{\normalfont{Datasets}}
Our benchmark dataset is sourced from literature and encompasses 1600 data entries. Each entry provides information on processing parameters, material characteristics, MAM process, machine type, post-processing conditions, and the orientation used for mechanical property testing. These details serve as input features, while mechanical properties such as yield strength, ultimate tensile strength, elongation at break, elastic modulus, Vickers and Rockwell hardness, as well as $R_z$ roughness, constitute the labels. These entries span diverse experiments conducted with various materials, processing parameters, AM machines, and post-processing techniques across a wide range of MAM processes. Subsequently, machine learning algorithms are employed to predict the mechanical properties of additively manufactured components using the mentioned input features, involving various regression tasks.
\vspace{3mm}

MAM processes are categorized based on the heat source (laser, electron beam, electric arc) and the method of raw material supply (powder or wire feed). Our dataset encompasses two main classes of MAM processes: powder bed fusion (PBF) and directed energy deposition (DED). Table \ref{tab:table1} lists different PBF and DED sub-processes in our benchmark dataset, and the standard terminologies which have been designated for each of them. For instance, PBF processes use either a laser or an electron beam to melt powder particles, classifying them into two sub-processes based on the heat source: laser powder bed fusion (L-PBF) and electron beam powder bed fusion (E-PBF).(Table \ref{tab:table1}) L-PBF process is also known and reported as selective laser melting (SLM), direct metal laser melting (DMLM), direct metal laser sintering (DMLS), and laser metal fusion (LMF) in the works that the dataset has been extracted from. Therefore, all of these commercial trade names have been reconciled into the standard terminology-- L-PBF. They have been commercialized by multiple manufacturers, and therefore, our benchmark machine dataset includes the mechanical properties of additive manufactured components fabricated by various AM machines such as EOS, SLM Solutions, Renishaw, Concept Laser, 3D Systems, Velo3D, and AddUp. It Also contains Arcam machines mechanical properties data, the company which has commercialized the E-PBF process.

\vspace{3mm}

\begin{table}[h!]
\begin{center}
\caption{MAM processes, sub-processes, and AM machine manufacturers in the benchmark dataset.}\vspace{2mm}
\label{tab:table1}
\scalebox{0.67}{
\begin{tabular}{ c |c c| c c c c}
\toprule [1pt]
 MAM Process &  PBF & &DED & &&  \\ [1ex]
&&&&&& \\ [1ex]
\toprule [2pt]
Feed-stock material &  Powder & Powder& Powder & Wire & Wire& Wire  \\ [1ex]
Heat source &  Laser & Electron Beam & Laser & Electron Beam & Electric Arc & Laser \\ [1ex]
\midrule [1pt]
Standard Terminology & L-PBF & E-PBF & L-DED & E-DED & Arc-DED & Wire-L-DED \\ [2ex]
\toprule [1pt]
Commercial MAM  & SLM & EBM &LENS & EBAM& WAAM& \\ [1ex]
Trade Name & DMLS && LMD && SMD &\\ [1ex]
& DMLM && DMD &&&\\ [1ex]
& LMF && DLD &&&\\ [1ex]
&&& LSF&&&\\ [1ex]
&&& Lf3 (FFF) && &\\ [1ex]
&&& DMP &&& \\ [1ex]
\toprule [1pt]
Manufacturer & EOS & Arcam &Optomec &Sciaky&  &\\[1ex]
& SLM solutions&&Markforged&&&  \\ [1ex]
&Renishaw&&Meltio&&& \\ [1ex]
&Concept Laser&&Desktop Metal&&& \\ [1ex]
&3D systems&&Trumpf &&& \\ [1ex]
& Velo3D &&&&& \\[1ex]
& AddUp &&&&& \\ [1ex]
\bottomrule
\end{tabular}
}
\end{center}
\end{table}

\begin{figure}\centering
\includegraphics[scale=0.085]{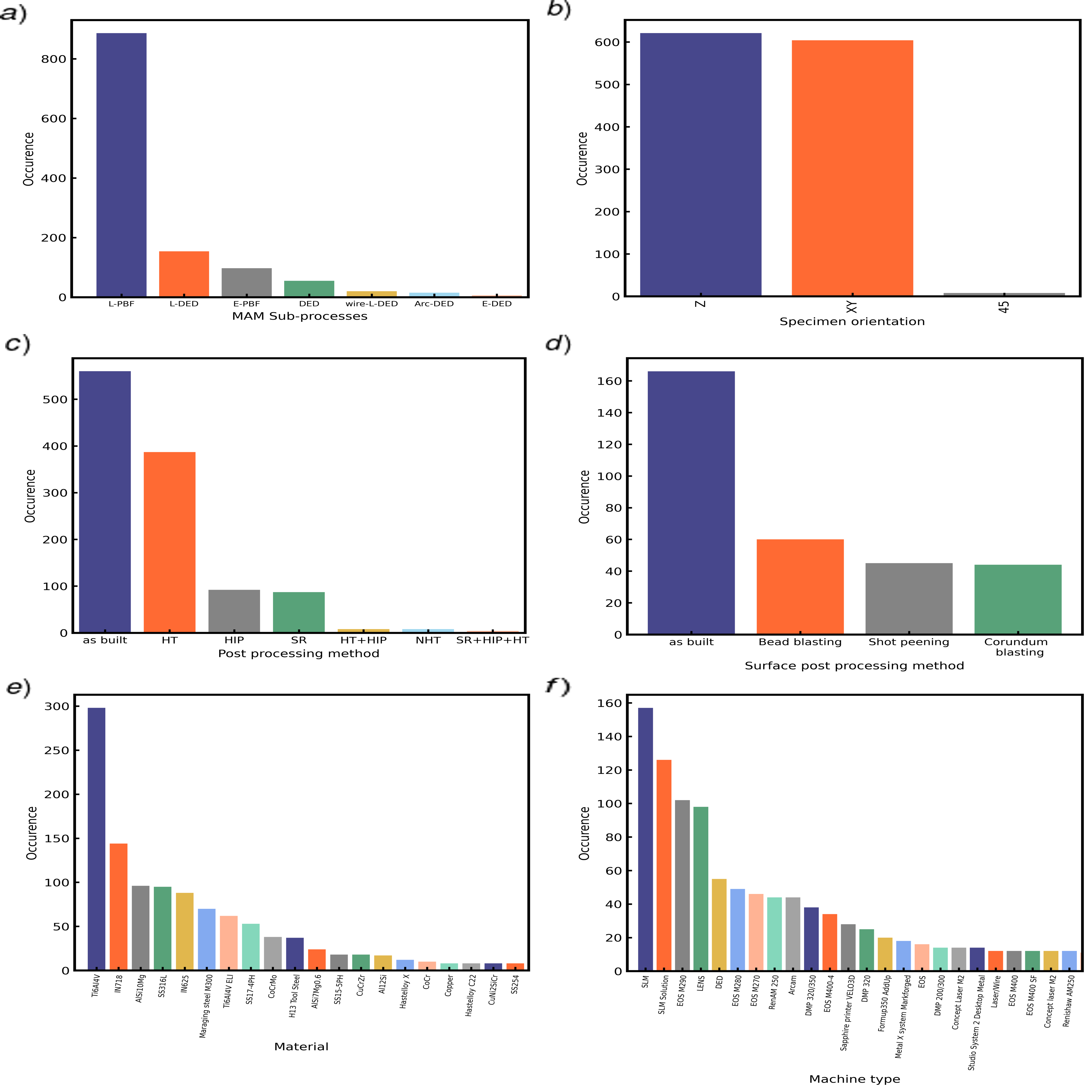}
\caption{a) Distribution of various MAM sub-processes within our benchmark dataset, b) Distribution of orientation within our benchmark dataset, c) Distribution of different post-processing methods and conditions within our dataset, d) Distribution of surface post-processing methods within our dataset, e) Occurrence of materials studied in our benchmark dataset, and f) Frequency of investigated additive manufacturing machines in our benchmark dataset.}
\label{fig:fig2}
\end{figure}

Additionally, DED processes are grouped based on the heat source (laser, electron beam, and electric arc), and material deposition method (powder or wire fed) resulting in four sub-categories: laser powder-fed directed energy deposition  (L-DED), Electron beam wire fed DED (E-DED), electric arc wire fed DED (arc-DED), and laser wire fed DED (wire-L-DED). (Table \ref{tab:table1}) The L-DED process employs a laser beam to meld deposited powder particles, creating a fully dense structure. This method goes by several names including laser engineering net shape (LENS), laser metal deposition (LMD), direct metal deposition (DMD), direct laser deposition (DLD), laser solid forming (LSF), laser freeform fabrication (LF3), and direct metal printing (DMP). Thus, the developed dataset covers a spectrum of DED processes including those offered by companies like Optomec, Markforged, Meltio, and Desktop Metal, among others. Additionally, it includes the E-DED process, known as electron beam additive manufacturing (EBAM), offered by Sciaky, along with arc-DED processes such as wire arc additive manufacturing (WAAM) and shaped metal deposition (SMD), as well as laser wire fed DED. Table \ref{tab:table1} provides an overview of the MAM processes, sub-processes, and the AM companies' machines included in our benchmark dataset. The occurrences of MAM sub-processes and their associated machine models within the dataset are depicted in Fig. \ref{fig:fig2}a and f.

\vspace{3mm}

We conducted an analysis on the mechanical characteristics of 68 distinct metals manufactured using MAM processes. (Fig. \ref{fig:fig2}e) Tables \ref{sec:sample:appendix:tab:tableA1} and \ref{sec:sample:appendix:tab:tableA2} present the chemical composition, density, specific heat, coefficient of thermal expansion, thermal conductivity, and melting temperature of these materials. Nearly 600 data points within the dataset provide details on the mechanical properties of printed metals in their as-built condition, without any post-processing methods. However, the remaining data points encompass a variety of post-processing techniques, including heat treatment (HT), hot isostatic pressing (HIP), and stress relieving (SR) (Fig. \ref{fig:fig2}c). Furthermore, the dataset on $R_z$ roughness includes information on various surface post-processing conditions such as as-built, bead blasting, shot peening, and corundum blasting (Fig. \ref{fig:fig2}d).

\vspace{3mm}

\begin{figure}\centering
\includegraphics[scale=0.06]{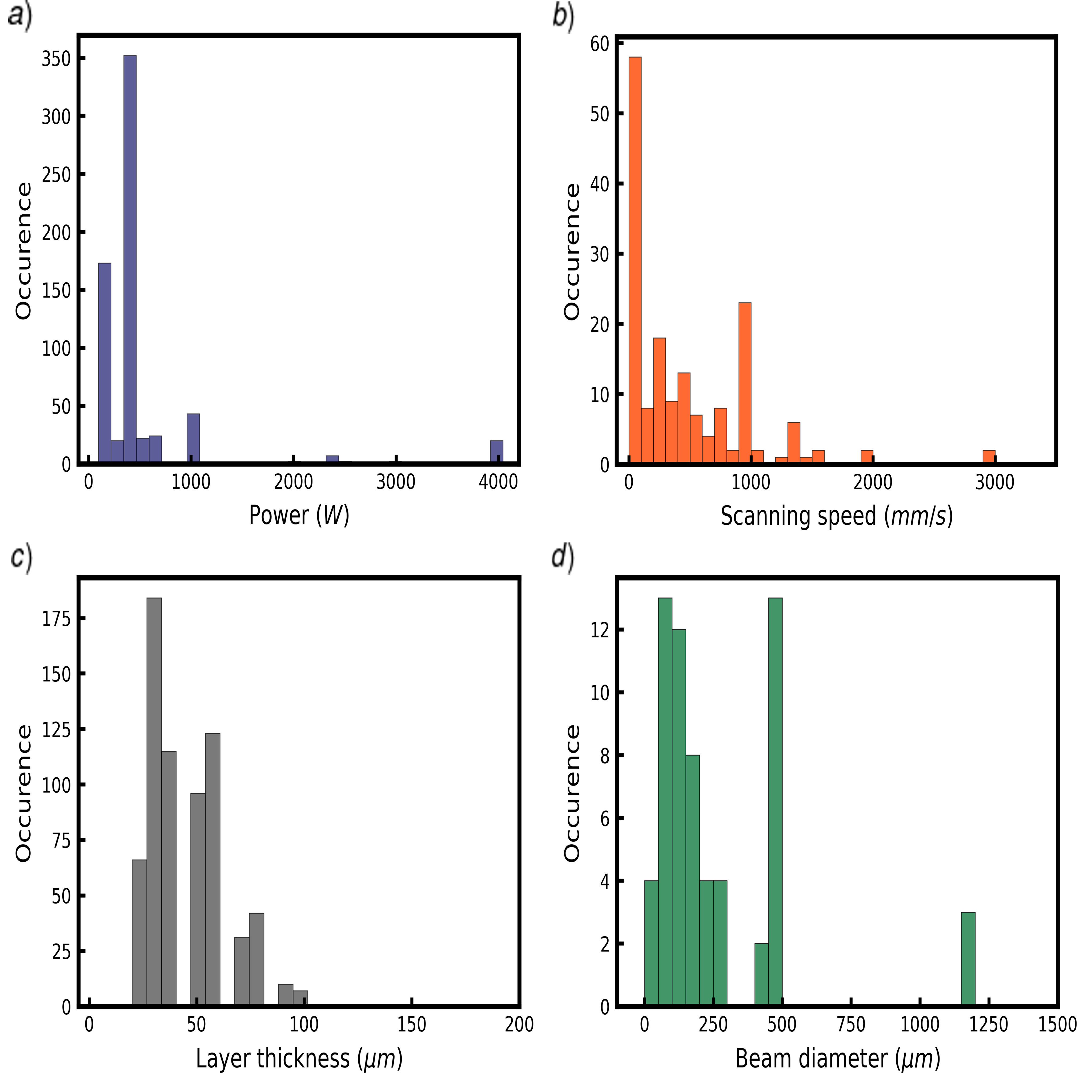}
\caption{The dataset processing parameters histograms and their distribution. a) Beam power histogram and occurrence, b) Scanning speed histogram and occurrence, c) Layer thickness histogram and occurrence, d) Beam diameter histogram and occurrence. }
\label{fig:fig3}
\end{figure}

The orientation in which parts are built significantly influences their mechanical properties in additive manufacturing processes, as highlighted in \cite{10.1115/1.4035420}. For instance, the mechanical characteristics of a printed component differ between horizontal and vertical orientations. Therefore, our dataset incorporates vertical, horizontal, and 45-degree orientations to accommodate the varied impact of orientations on mechanical properties (Fig. \ref{fig:fig2}b). Furthermore, Fig \ref{fig:fig3} illustrates the distribution of processing parameters—such as beam power, scanning speed, layer thickness, and beam diameter—within the dataset. It is noteworthy that during the training of machine learning models, only data points containing all the requisite input features are selected for inclusion in the dataset. Our benchmark currently comprises 1600 data samples on MAM mechanical properties and is anticipated to expand continuously. Hence, we welcome contributions from other public data collections.

\subsection{\normalfont{Featurization}}
During the machine learning training phase, it is crucial to select an appropriate set of features to input into the ML models to ensure reliable prediction performance. The foundational featurization method adopted in our benchmark is outlined in Table \ref{tab:table2}. This includes essential processing parameters such as beam power and layer thickness, consistently reported across the majority of references for both powder bed fusion (PBF) and directed energy deposition (DED) processes. Additionally, fundamental material properties like density, specific heat, thermal conductivity, coefficient of thermal expansion (CTE), and melting temperature serve as core quantitative features.

\vspace{3mm}

\begin{table}[ht!]
\begin{center}
\caption{Baseline featurization employed in our benchmark ML models.}\vspace{2mm}
\label{tab:table2}
\scalebox{0.8}
{
\begin{tabular}{c| c | c }
\toprule [1pt]
Processing parameters &  Material properties & Categorical features  \\ [1ex]
\midrule [2pt]
Beam Power   &  Density & Material \\ [1ex]
Layer thickness & Melting point & Machine type  \\ [1ex]
& Thermal conductivity &  Specimen orientation\\ [1ex]
& Specific heat &  Post-processing condition \\ [1ex]
& Coefficient of thermal expansion &  MAM process \\ [1ex]
&  &  MAM sub-process \\ [1ex]
\bottomrule
\end{tabular}
}
\end{center}
\end{table}

Furthermore, the baseline featurization encompasses six categorical features, namely Material, AM machine type, MAM process and sub-process, post-processing condition, and specimen orientation. These categorical features undergo featurization using one-hot encoding. One-hot encoding transforms categorical features into numeric categories suitable for processing by ML algorithms. It generates a one-hot vector of size $1 \times n$ for each category. Consequently, for a data point falling within the $k$-th category, it assigns a value of $1$ to the $k$-th index while assigning a value of $0$ to the other $n-1$ indices of the one-hot vector. This encoding method effectively captures distinctions between different MAM processes, sub-processes, machines, post-processing methods, and build orientations, facilitating prediction tasks. To ensure uniform information provision to the ML models across specific feature combinations, data points lacking one or more features are excluded from the training process. This step is essential because not all collected data samples contain all the features employed in the featurization process.

\subsubsection{\normalfont Material featurization} 

In MAM processes, alloys typically comprise multiple elements with varying concentrations. Within our benchmark, besides featurizing materials with their respective thermal properties (refer to Table \ref{tab:table2}), they have also undergone featurization using one-hot encoding. Additionally, alternative methods have been explored, including material chemical compositions (wt\%) and elemental featurization.

\vspace{3mm}

The initial approach involves incorporating the mass percentages of individual elements within an alloy into the baseline feature representation. Alternatively, the elemental featurization integrates the elemental characteristics into the baseline feature set for predicting mechanical properties. The considered elemental properties are detailed in Table \ref{tab:table3}. The dataset's elemental properties were gathered using the python Mendeleev package \cite{mendeleev}. Our dataset comprises 20 chemical elements, some of which are constituents of various alloys. To account for the influence of each element's relative concentration on the overall alloy properties, a linear mixture rule \cite{CARRUTHERS1988351} is employed:

\begin{equation}
x_i = \sum_{j}^{N} x_{ij}  a_j 
\end{equation}
where $x_i$ is the $i^{\text{th}}$ feature, $x_{ij}$ is the $i^{\text{th}}$ feature of the $j^{\text{th}}$ component, and $a_i$ is the $j^{\text{th}}$ element fraction. 

\begin{table}[ht!]
\begin{center}
\caption{Properties utilized in our elemental featurization, a method for featurizing materials.}\vspace{2mm}
\label{tab:table3}
\scalebox{0.85}  
{
\begin{tabular}{c c }
\toprule [1pt]
Elemental featurization&  \\ [1ex]
\midrule [2pt]
Atomic number & Heat of fusion \\ [1ex]
Atomic volume & Electron affinity \\ [1ex]
Ionization energy&   \\ [1ex]
\bottomrule
\end{tabular}
}
\end{center}
\end{table}

\subsection{\normalfont{Data splitting and Evaluation Metrics}}

Table \ref{tab:table4} provides an overview of the datasets within our collection, including details on prediction tasks, the split method employed, and the evaluation metrics utilized. The heterogeneous nature of the quantitative input features within the dataset, characterized by various measurement units, can detrimentally affect the predictive accuracy of machine learning models such as Support Vector Machine (SVM), as well as Lasso and Ridge linear regression. Consequently, the dataset undergoes standardization, resulting in a mean of zero and a unit standard deviation for all features, thereby ensuring uniform scaling. The standardization process is implemented according to the following equation:

\begin{equation}
x_{Normalized} = \frac{x - \bar{x}}{\sigma}
\end{equation}
Where $\bar{x}$ signifies the average of the input parameters, and $\sigma$ stands for their standard deviation. 
\vspace{3mm}
 
To establish robust ML models and assess their performance on unseen data, it is necessary to partition the dataset into training and testing subsets. This ensures that models are trained on the training set during the learning phase, while the test set is exclusively reserved for evaluating the prediction accuracy of the trained machine learning algorithms. Accordingly, the dataset was initially randomly shuffled, and a 5-fold cross-validation approach was adopted, allowing for the examination of five distinct training and test sets. In the 5-fold cross-validation method, the dataset is divided into five folds. During each iteration, one of these folds is withheld to serve as the test set for evaluating the model, while the remaining four folds are used for model training. This process is iterated five times, ensuring that each fold is utilized as the test set once. Consequently, all benchmark results presented in the subsequent section represent averages over five iterations. The mean absolute error (MAE) and the coefficient of determination ($R^{2}$) are also employed to assess the prediction performance of the implemented machine learning models (refer to Table \ref{tab:table4}).

\begin{table}[ht]
\begin{center}
\caption{Our benchmark dataset details. (Tasks, splitting method, and metrics).}\vspace{2mm}
\label{tab:table4}
\scalebox{0.85}{
\begin{tabular}{c| c| c | c| c}
\toprule [1pt]
Category & Dataset   & Tasks     & Rec - split   &   Rec - metric \\
\midrule [2pt]
Yield strength  & 1218 &  &  & \\ [1ex]
Ultimate tensile strength  & 1244  &  &  &\\ [1ex]
Elastic modulus  & 432 &  &  & \\ [1ex]
Elongation at break & 1198 & Regression & Random & $R^2$ - MAE \\ [1ex]
Hardness (Vickers) & 293 &  &  & \\ [1ex]
Hardness (Rockwell) & 230 &  &  & \\ [1ex]
Roughness Rz & 218 &  &  &  \\ [1ex]
\bottomrule
\end{tabular} 
}
\end{center}
\end{table}

\subsection{\normalfont{Models}}
In this section, we will showcase various machine learning models applied to the benchmark dataset, highlighting their predictive performance in the subsequent results section.

\subsubsection{\normalfont Random Forest 'RF'}
The Random Forest (RF) algorithm, rooted in ensemble learning principles, is a robust technique for regression tasks \cite{RF}. RF employs a combination of bagging and bootstrap aggregating (boostraping) methods. In the bagging process, RF creates multiple decision trees in parallel, each trained on a randomly sampled subset of the dataset with replacement. This technique introduces diversity among the trees, reducing the risk of overfitting and enhancing generalization. Bootstrap aggregating, or bootstrapping, involves sampling with replacement from the original dataset to generate multiple subsets of data for training. 

\vspace{3mm}

This process ensures that each decision tree in the RF ensemble is exposed to different instances of the dataset, thereby improving the model's robustness and reducing variance. During training, each decision tree in the RF ensemble independently learns to predict the target variable. For regression tasks, RF aggregates the predictions of all trees to produce a final prediction, typically by averaging their outputs. This ensemble approach often yields superior predictive performance compared to individual decision trees. One notable advantage of RF is its ability to handle high-dimensional datasets with a large number of features, as it automatically selects a subset of features for each split, thereby reducing the risk of overfitting and enhancing computational efficiency.

\subsubsection{\normalfont Gradient Boosting Trees 'GB'}
Gradient Boosting (GB) is another ensemble learning technique that utilizes decision trees' outputs for prediction \cite{GB}. In contrast to the Random Forest algorithm, GB employs shallow decision trees known as weak learners and combines their results throughout the training process. GB adopts a sequential approach, building one tree at a time. Each new tree focuses on correcting the errors made by the previously trained decision tree. By iteratively improving upon the mistakes of its predecessors, GB gradually constructs a strong predictive model. 

\vspace{3mm}

One key advantage of GB is its ability to capture complex relationships within the data, as the model iteratively learns from its errors and refines its predictions. Additionally, GB is less prone to overfitting compared to deep decision trees, making it suitable for a wide range of regression tasks. However, it is important to note that Gradient Boosting typically requires more computational resources and tuning compared to Random Forest, as it involves a sequential learning process. Despite this, GB often achieves superior predictive performance, especially on structured datasets with intricate patterns.

\subsubsection{\normalfont Support Vector Regressor ('SVR')}
Support Vector Machines (SVMs) are powerful supervised learning algorithms commonly employed for regression tasks. Support Vector Regression (SVR) is a variant of SVM tailored for regression analysis. SVR allows us to determine how much error is bearable in our model, and operates by fitting a hyperplane to the data points in the feature space. This hyperplane serves as the decision boundary, with support vectors delineating the boundary at a distance such that data points fall within its margins. By optimizing the hyperplane's position and margin width, SVR aims to minimize prediction errors while maximizing the margin between the hyperplane and the support vectors \cite{svm}.

\vspace{3mm}

One notable feature of SVR is its flexibility in handling non-linear relationships between input features and the target variable. This is achieved through the use of kernel functions, which transform the input features into a higher-dimensional space where a linear hyperplane can effectively separate the data points. Despite its effectiveness, SVR's performance heavily relies on proper selection of hyperparameters and kernel functions. Additionally, SVR may not be well-suited for datasets with a large number of features or instances due to its computational complexity. However, when properly tuned and applied to appropriate datasets, SVR can yield accurate regression predictions with robust generalization capabilities.

\subsubsection{\normalfont Gaussian Process Regressor ('GPR')}
Gaussian Process Regressors (GPR) are probabilistic, Bayesian models that offer a departure from the conventional output approach of many other machine learning models. Instead of providing precise values for parameters, GPRs generate a probability distribution encompassing all potential output values \cite{Rasmussen2004}. In scenarios requiring uncertainty quantification, GPRs prove highly advantageous. Their ability to furnish a complete probability distribution allows for discerning the confidence level associated with each prediction. This capability facilitates informed decision-making, especially in uncertain environments such as anomaly detection, where assessing prediction uncertainty is critical for devising effective risk mitigation strategies.

\vspace{3mm}

Furthermore, GPRs inherently capture data uncertainty, rendering them resilient to noisy or sparse datasets. This capacity to model uncertainty enhances prediction reliability, particularly when confronted with limited or incomplete data. However, it's noteworthy that GPRs may entail computational complexity, especially as dataset sizes increase. Additionally, meticulous selection of hyperparameters and kernel functions is pivotal for optimizing GPR performance. Nonetheless, owing to their probabilistic nature, GPRs remain invaluable tools across various machine learning domains, particularly those emphasizing uncertainty quantification.

\subsubsection{\normalfont Lasso Linear Regression 'Lasso'}
The Least Absolute Shrinkage and Selection Operator (LASSO) serves as a regularized variant of linear regression, designed to mitigate overfitting by employing $L_1$ regularization techniques \cite{lasso}. Overfitting occurs when the model becomes excessively complex, fitting training data points too closely and yielding a low training error but a significantly higher test error. By integrating an $L_1$ penalty term into the cost function, Lasso regression effectively nullifies irrelevant parameters, setting them to zero. Consequently, this reduces model complexity and curbs overfitting tendencies. Lasso regression's distinctive feature lies in its capability to perform variable selection, effectively excluding irrelevant features from the model. This characteristic is particularly beneficial in scenarios where the dataset encompasses numerous potentially redundant or insignificant features, streamlining the model and enhancing interpretability.
\subsubsection{\normalfont Ridge Linear Regression 'Ridge'}
Ridge regression, akin to Lasso, offers a regularization mechanism for linear regression, albeit through $L_2$ regularization \cite{Ridge}. This method aims to alleviate overfitting by imposing constraints on parameter values, thereby minimizing their impact on the trained model. Unlike Lasso, which may entirely eliminate some features, Ridge regression reduces the influence of all features, proportionally shrinking their coefficients. This reduction in parameter magnitudes leads to a smoother model and helps mitigate the adverse effects of overfitting, particularly in scenarios where the dataset contains multicollinear features.

\subsubsection{\normalfont Neural Network 'NN'}
Neural Networks (NN) represent a subset of machine learning and serve as the foundation of deep learning methodologies \cite{NN}. Drawing inspiration from the biological structure of the human brain, NNs comprise interconnected neurons that communicate with one another. These networks typically feature an input layer, one or more hidden layers, and an output layer. During training, NNs undergo an iterative process where the initially random weights are adjusted via backpropagation. This process continues until the model converges to a global minimum, minimizing the loss function. By iteratively fine-tuning the weights based on observed errors, NNs can effectively learn complex patterns and relationships within the data.

\vspace{3mm}

NNs have gained significant traction due to their ability to handle large and diverse datasets, as well as their capability to learn hierarchical representations of data. They excel in tasks such as image and speech recognition, natural language processing, and many others. However, training deep neural networks can be computationally intensive and may require substantial amounts of labeled data. Nonetheless, NNs remain at the forefront of many cutting-edge machine learning applications, driving advancements in various fields.

\subsubsection{\normalfont XGBoost} 
XGBoost, short for Extreme Gradient Boosting, represents an enhanced variant of Gradient Boosting algorithms, featuring regularization techniques like L1 and L2 regularization for improved generalization capabilities \cite{Chen2016}. In contrast to traditional gradient boosted trees, XGBoost offers superior speed, high performance, and the ability to conduct parallel tree boosting. One notable advantage of XGBoost lies in its efficiency, making it well-suited for large-scale datasets and computationally intensive tasks. Moreover, XGBoost has gained popularity for its robustness and versatility across various machine learning applications.

\vspace{3mm}

In our paper, XGBoost is also employed for feature importance analysis and SHAP (SHapley Additive exPlanations) analysis. These techniques enable practitioners to gain insights into the relative importance of input features and understand the model's decision-making process. This dual functionality enhances the interpretability of XGBoost and makes it a valuable tool for both predictive modeling and interpretability in machine learning tasks.

\subsection{\normalfont Hyperparameter optimization}

Within each machine learning (ML) model lie two types of parameters: model parameters and hyperparameters. Model parameters are the internal variables learned from the training data, whereas hyperparameters are adjustable settings that govern the learning process, such as learning rate, regularization strength, and choice of optimizer. Therefore, selecting optimal values for these hyperparameters is crucial for achieving high-performance ML models and refining the learned parameters during training.

\vspace{3mm}

Various methods exist for optimizing ML hyperparameters, including Grid Search, Random Search, and Bayesian Optimization. In Grid Search, the algorithm trains and evaluates the model on every possible combination of predefined hyperparameter values. Once completed, the model achieving the highest accuracy is deemed the best, and its hyperparameters are selected as optimized. Conversely, Random Search trains the model on randomly selected hyperparameter configurations. While Random Search can be more efficient than Grid Search \cite{JMLR:v13:bergstra12a}, both methods lack information from past evaluations, potentially leading to significant time spent evaluating suboptimal hyperparameters.

\vspace{3mm}

In contrast, Bayesian approaches leverage past evaluation results to iteratively construct a probabilistic model of the objective function. This surrogate model guides the selection of hyperparameters, which are then evaluated on the actual objective function. With each iteration, the surrogate model is refined based on new results, enabling Bayesian Optimization to make more informed decisions compared to random and grid search methods. \cite{bergstra2011algorithms}

\vspace{3mm}

To implement Bayesian Optimization, the Hyperopt Python package, which employs the Tree-structured Parzen Estimator (TPE) algorithm was utilized to optimize the validation $R^2$ score objective function \cite{hyperopt}. The exploration of hyperparameters and their respective ranges for regression tasks such as yield strength, ultimate tensile strength, and elastic modulus is detailed in Table \ref{tab:table5}. Additionally, Table \ref{sec:sample:appendix:tab:tableA3} provides the optimal hyperparameter values for elongation, Vickers and Rockwell hardness, and $R_z$ surface roughness.

\vspace{3mm}

\begin{table}[ht!]
\begin{center}
\caption{Hyperparameters and their range studied in our benchmark ML models.}\vspace{2mm}
\label{tab:table5}
\scalebox{0.63}{
\begin{tabular}{c c c c c}
\toprule [1pt]
Regression ML Task & {Models} & Hyperparameters & Value &  Range studied   \\ [1ex]
\midrule [2pt]
Yield strength & Random Forest & n\_estimators &	382 &1-500	\\ [2ex]
 & Support Vector Regressor & C & 698	&1-1000	\\ [0.8ex]
 &  & kernel & 'poly'	&['linear', 'poly', 'rbf', 'sigmoid']	\\ [0.8ex]
  &  & degree & 2	&[2,3,4]	\\[2ex]
 & Gradient boosting &n\_estimators & 500 &1-500	\\ [2ex]
 & Neural network & number of neurons &(128,256,32)	&[32, 64, 128, 256, 512]\\ [0.8ex]
 &  & alpha &0.04943663345976882	& 1e-7 -  1e-1\\ [2ex]
\midrule [0.5pt]
Ultimate tensile  & Random forest & n\_estimators &	500 &1-500\\ [2ex]
strength & Support vector classifier & C & 982	&1-1000	\\ [0.8ex]
 &  & kernel & 'rbf'	&['linear', 'poly', 'rbf', 'sigmoid'] \\ [2ex]
 & Gradient boosting &n\_estimators & 500 &1-500	\\ [2ex]
  &  Neural network & number of neurons &(256,256,512)	&[32, 64, 128, 256, 512]\\ [0.8ex]
 &  & alpha &0.04030067181655384	& 1e-7 -  1e-1\\ [1ex]
\midrule [0.5pt]
Elastic modulus & Random Forest & n\_estimators &	43 &1-500	\\ [2ex]
 & Support Vector Regressor & C & 43	&1-1000	\\ [0.8ex]
 &  & kernel & 'rbf'	&['linear', 'poly', 'rbf', 'sigmoid']	\\ [2ex]
 & Gradient boosting &n\_estimators & 316 &1-500	\\ [2ex]
 & Neural network & number of neurons &(32,512,256)	&[32, 64, 128, 256, 512]\\ [0.8ex]
 &  & alpha &0.041329503263237435	& 1e-7 -  1e-1\\ [2ex]
\bottomrule 
\end{tabular} 
}
\end{center}
\end{table}

In the case of Random Forest and Gradient Boosting algorithms, the hyperparameter 'n-estimators', which signifies the number of decision trees in the forest, undergoes optimization. The 'n-estimators' parameter controls the complexity of the ensemble model and influences its predictive power. While a higher number of decision trees tends to enhance the performance of these models by reducing overfitting and increasing robustness to noise in the data, it also leads to increased computation time and cost. Each additional tree requires more memory and computational resources during both training and inference. Therefore, determining an optimal value for the number of decision trees holds paramount importance in balancing model performance with computational efficiency \cite{probst2019hyperparameters}.

\vspace{3mm}

In our exploration of the Support Vector Machine (SVM) algorithm, our investigation encompassed four distinct SVM kernels: linear, polynomial, radial basis function (rbf), and sigmoid. While Linear SVR involves fitting linear support vectors over the data points, situations may arise where linear hyperplanes prove insufficient in capturing the data distribution effectively. In such cases, non-linear kernels such as polynomial, rbf, and sigmoid were leveraged to achieve better model performance \cite{scholkopf2002learning}. Moreover, optimization of the regularization hyperparameter, represented as $C$, was explored. This parameter, which governs the balance between achieving a low training error and minimizing model complexity, plays a critical role in refining SVM models and mitigating overfitting \cite{DUAN200341}. By thoroughly exploring these aspects, the aim was to develop SVM models that strike a fine balance between flexibility and generalization ability, ensuring robust performance across diverse datasets.

\vspace{3mm}
In the optimization process of the Neural Network algorithm, our primary focus was on identifying the optimal number of neurons in each layer. It is crucial to strike a balance between increasing model complexity with more neurons and avoiding overfitting. Additionally, the optimal value for alpha, which serves as a regularization parameter, was examined. Alpha imposes constraints on model weights using the $L_2$ norm, effectively countering overfitting tendencies by penalizing large weights. This comprehensive approach ensures not only the enhancement of model performance but also guards against the pitfalls of overfitting, thereby promoting robust and reliable neural network models.

\vspace{3mm}

\vspace{3mm}
\section{Results and discussion}
In this section, we delve into the predictive capabilities of the machine learning (ML) models employed on the benchmark dataset. It is important to note that all presented benchmark results are averaged across five iterations, with their standard deviations depicted as error bars. Furthermore, the correlation matrix of the mechanical properties within the dataset is analyzed. Additionally, a drop-column feature importance study is conducted to discern the most crucial and influential features in each dataset and task. Furthermore, the impact of varying the size of the training set on the models' performance is explored. Moreover, data-driven explicit models for the mechanical properties are identified. Detailed findings will be discussed in the subsequent sections.

\vspace{3mm}

\subsection{\normalfont ML models results}

The results, depicted in Fig. \ref{fig:fig4}, showcase the effectiveness of various machine learning models, including the Random Forest Regressor, Gaussian Process Regressor, Support Vector Regressor, Ridge Linear Regressor, Lasso Linear Regressor, Gradient Boosting, and Neural Network, in predicting yield strength, ultimate tensile strength, Elastic Modulus, and the elongation at break. These results are evaluated based on their respective accuracy, measured by $R^2$ (Fig. \ref{fig:fig4} left column), and Mean Absolute Error (MAE) (Fig. \ref{fig:fig4} right column). Initially, the models were trained using baseline featurization, as detailed in Table \ref{tab:table2}, and these features were inputted into the ML models (Fig. \ref{fig:fig4}, baseline model). Subsequently, additional featurization methods were explored to enhance the prediction accuracy. In addition to thermal properties and one-hot encoding of materials included in the baseline model, the chemical composition of materials in terms of weight percentage (Fig. \ref{fig:fig4}, baseline + chemical composition mass percentage model) and elemental featurization (Fig. \ref{fig:fig4}, baseline + elemental featurization model) were examined. These supplementary featurization techniques yielded improvements in prediction accuracy for certain ML models.

\vspace{3mm}

\begin{figure}\centering
\includegraphics[scale=0.05]{Figures/figure_4.jpg}
\caption{Evaluation of benchmark performances for yield strength, ultimate tensile strength, Elastic Modulus, and Elongation at break: Various machine learning models including 'Random Forest', 'Gaussian Process Regressor', 'Support Vector Regressor', 'Ridge Linear Regressor', 'Lasso Linear Regressor', 'Gradient Boosting', and 'Neural Network' are assessed based on their $R^2$ score (left column) and Mean Absolute Error (MAE) (right column). a) $R^2$ accuracy and MAE results for predicting yield strength, b)$R^2$ accuracy and MAE results for predicting ultimate tensile strength, c) $R^2$ accuracy and MAE results for predicting Elastic Modulus, d) $R^2$ accuracy and MAE results for predicting Elongation at break. (Note: Higher $R^2$ scores and lower MAE values indicate superior performance.)}
\label{fig:fig4}
\end{figure}
As depicted in Fig. \ref{fig:fig4}a, for the yield strength, the best results belongs to the last model for the Random forest algorithm with 95.77\% accuracy, and an MAE of 53.97 $MPa$. (Fig. \ref{fig:fig4}a) In addition, for the ultimate tensile strength (Fig. \ref{fig:fig4}b), the best results were achieved with the fourth model which yields an $R^2$ accuracy of 96.94\% and an MAE of 47.95 $MPa$. Furthermore, following drop-column feature importance analysis and the selection of features that positively impact prediction performance for each task, along with hyperparameter optimization, there was a significant improvement in results (depicted as gray bars in Fig.\ref{fig:fig4}). As shown in Fig.\ref{fig:fig4}a, in the case of yield strength prediction, the most favorable outcomes were observed with the final model utilizing the Random Forest algorithm, achieving an accuracy of 95.77\% and an MAE of 53.97 $MPa$. Similarly, for ultimate tensile strength prediction (Fig.\ref{fig:fig4}b), the best performance was attained by the fourth model, which achieved an $R^2$ accuracy of 96.94\% and an MAE of 47.95 $MPa$.

\vspace{3mm}

For predicting Elastic modulus, the most promising outcome was achieved through feature selection and hyperparameter tuning using the Random Forest algorithm, yielding an accuracy of 94.27\% and an MAE of 6.63 $MPa$. (Fig. \ref{fig:fig4}c) Moreover, for predicting elongation at break, the Random Forest model exhibited the highest $R^2$ accuracy of 88.66\%, accompanied by an MAE of 2.89. (Fig. \ref{fig:fig4}d)

\vspace{3mm}

The predictive accuracy ($R^2$) and MAE results of the ML models on the Vickers and Rockwell hardness as well as $R_z$ roughness prediction tasks have also been assessed (Fig. \ref{fig:fig5}). our analysis included elemental featurization and the selection of the most influential features combined with hyperparameter optimization (Fig. \ref{fig:fig5}). In the case of Vickers hardness prediction, the third model employing the Random Forest algorithm yielded the highest $R^2$ accuracy of 99.62\%, while the Gradient Boosting model achieved the lowest MAE of 15.58 HV (Fig. \ref{fig:fig5}a). Moreover, for Rockwell hardness prediction, the Gradient Boosting model demonstrated superior performance with an accuracy of 96.05\% and an MAE of 2.20 HRC (Fig. \ref{fig:fig5}b). Additionally, for the $R_z$ roughness, the neural network model outperformed others with a baseline accuracy of 90.07\% and an MAE of 4.69 $\mu m$.

\begin{figure}\centering
\includegraphics[scale=0.05]{Figures/figure_5.jpg}
\caption{Evaluation of benchmark performances for the Vickers hardness, the Rockwell hardness, and $R_z$ Roughness; Various machine learning models including 'Random Forest', 'Gaussian Process Regressor','Support Vector Regressor', 'Ridge Linear Regressor', 'Lasso Linear Regressor', Gradient Boosting', 'Neural Network' are assessed based on their $R^2$ score (left column) and Mean Absolute Error (MAE) (right column). a) $R^2$ accuracy and MAE results for predicting Vickers hardness, b) $R^2$ accuracy and MAE results for predicting the Rockwell hardness, c) $R^2$ accuracy and MAE results for predicting $R_z$ Roughness. (Note: Higher $R^2$ scores and lower MAE values indicate superior performance.)}
\label{fig:fig5}
\end{figure}

\begin{table}[ht]
\begin{center}
\caption{Overview of machine learning model performances in regression tasks (Yield strength, Ultimate tensile strength, Elastic modulus, Elongation at break, Hardness, and mean roughness depth) (test subset). (Optimal performance values are presented for each category using two distinct metrics.)}\vspace{1mm}
\label{tab:table6}
\scalebox{0.8}{
\begin{tabular}{c| c c  c c c}
\toprule [1pt]
Category & Metric            & Best performance  & Value  \\
\midrule [2pt]
Yield strength  & $R^2$ &  Random forests  & 0.9580 \\ [1ex]
  & MAE &  Random forests & 53.55 $MPa$ \\ [1ex]

Ultimate tensile strength  & $R^2$ & Random forests & 0.9694\\ [1ex]
  & MAE  & Random forests & 47.95 $MPa$ \\ [1ex]

Elastic modulus  & $R^2$ & Random forests  & 0.9427 \\ [1ex]
  & MAE  & Gradient Boosting  & 6.63 $GPa$ \\ [1ex]

Elongation at break & $R^2$   & Random forests & 0.8865 \\ [1ex]
 & MAE & Random forests & 2.89 \% \\ [1ex]

Hardness (Vickers) & $R^2$  & Random forests & 0.9381 \\ [1ex]
 & MAE & Random forests & 18.58 $HV$ \\ [1ex]
 
Hardness (Rockwell) & $R^2$  & Gradient Boosting & 0.9624 \\ [1ex]
 & MAE & Gradient Boosting & 1.92 $HRC$ \\ [1ex]
 
Mean Roughness depth ($R_z$) & $R^2$  & Neural network & 0.9007 \\ [1ex]
 & MAE & Neural network & 4.69 $\mu m$ \\ [1ex]
\bottomrule
\end{tabular} 
}
\end{center}
\end{table}

\vspace{3mm}

Table \ref{tab:table6} presents an overview of the top-performing ML models in the regression tasks. It is evident from the table that ensemble learning techniques such as Random Forest and Gradient Boosting, along with Neural Networks, exhibit superior performance compared to other algorithms utilized in these regression tasks. These models excel in providing reliable predictions, especially when dealing with large datasets characterized by complex, nonlinear relationships among numerous features. Random Forest, in particular, is well-suited for datasets with disparate scales and can effectively handle both categorical and numerical features, making it more robust against outliers and noisy data compared to Gradient Boosting. Conversely, linear regression models like Ridge and Lasso struggle to perform well on our dataset due to its nonlinear nature and potential presence of outliers and noise. Moreover, these models are typically more suitable for datasets exhibiting multicollinearity, which is not the case with our collected dataset. Although Gaussian Process models can accommodate nonlinear data, they are less practical in high-dimensional settings due to computational constraints \cite{liu2020gaussian}. Additionally, SVR showed satisfactory performance on the dataset.

\vspace{3mm}

Fig. \ref{fig:fig6} depicts the predicted values of the examined mechanical properties contrasted with their actual values in the test set, utilizing the ML models demonstrating the highest prediction accuracies. Achieving a satisfactory prediction performance, denoted by an $R^2$ value close to 1, implies that the predicted values for each data point closely align with the actual values, essentially lying on the diagonal line ($y=x$). Consequently, the results appear highly reliable, as evidenced by the data points clustering closely around the $y=x$ line, accompanied by narrow confidence bands.

\begin{figure}\centering
\includegraphics[scale=0.085]{Figures/figure_6.jpg}
\caption{Comparison between predicted and actual values plotted for the top-performing model in: a) Yield strength, b) Ultimate tensile strength, c) Elastic modulus, d) Elongation at break, e) Vickers hardness, f) Rockwell hardness, and g) $R_z$ Roughness.}
\label{fig:fig6}
\end{figure}

\subsection{\normalfont Correlation matrix of mechanical properties}

\begin{figure}\centering
\includegraphics[scale=0.062]{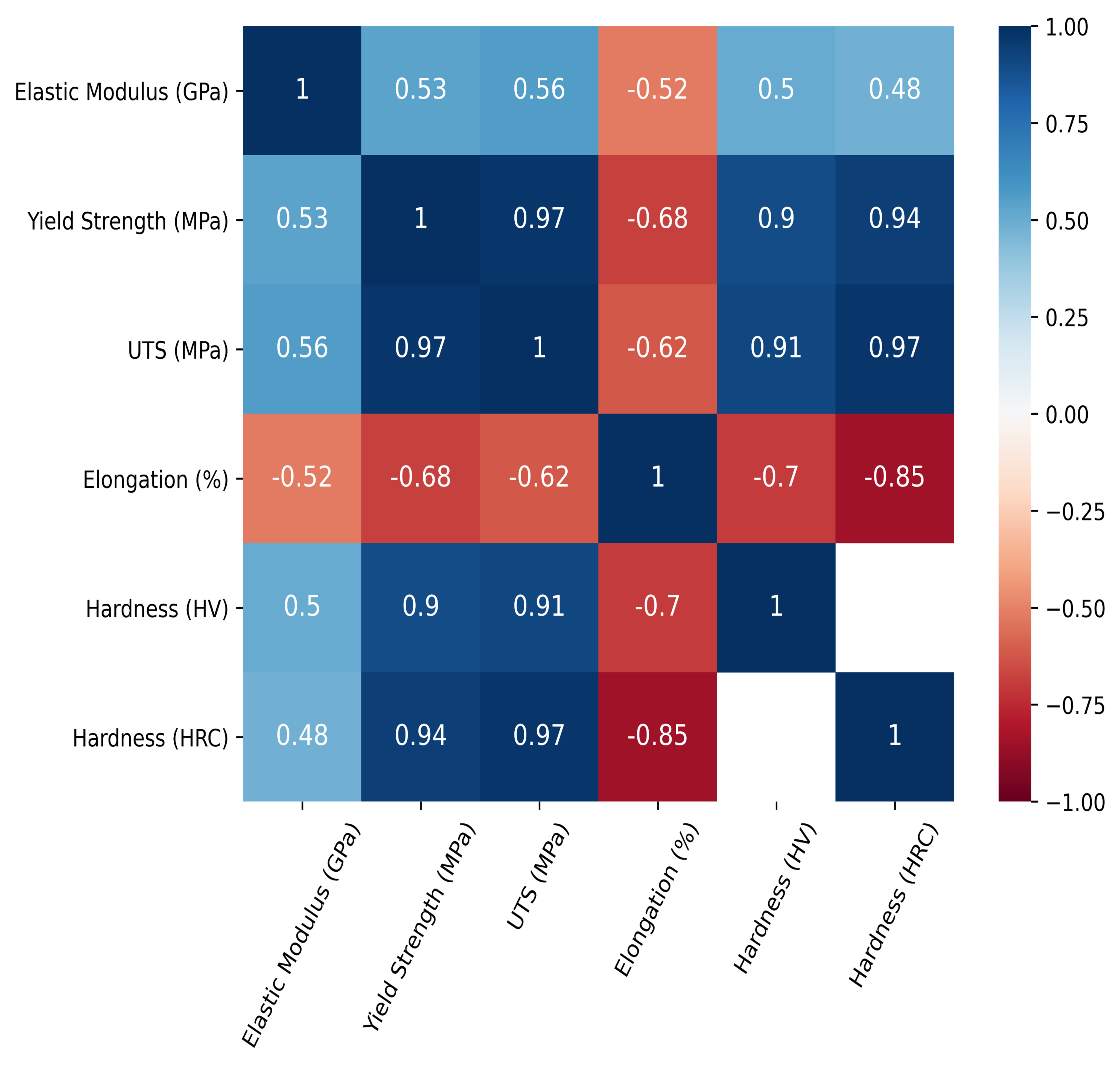}
\caption{Matrix showing the Pearson correlation among mechanical properties in the benchmark dataset of Inconel 718.}
\label{fig:fig7}
\end{figure}

The Pearson correlation matrix of the mechanical properties within our benchmark dataset for Inconel 718 is illustrated (Figure \ref{fig:fig7}). This matrix serves as a robust tool for summarizing a comprehensive dataset, displaying correlation coefficients and patterns among all variables. Notably, yield strength, ultimate tensile strength, and elastic modulus exhibit positive correlations with each other, while displaying negative correlations with elongation. Furthermore, hardness—both Vickers and Rockwell— exhibits strong positive correlations with material strength characteristics, yet displays negative correlations with elongation. These findings align closely with the well-established concept of the strength-ductility trade-off in materials science, wherein enhancing material strength inherently leads to a reduction in ductility \cite{wei2014evading}.

\vspace{3mm}

\begin{figure}\centering
\includegraphics[scale=0.065]{Figures/figure_9.jpg}
\caption{Feature importance analysis using the drop-column method for: a) yield strength, b) ultimate tensile strength, c) elastic modulus, d) elongation at break, e) Vickers hardness, f) Rockwell hardness, and g) $R_z$ roughness.}
\label{fig:fig8}
\end{figure}
\subsection{\normalfont Drop-column feature importance}

In order to identify the most crucial and decisive features in predicting each mechanical property, a drop-column feature importance analysis—considered one of the most accurate methods for feature importance—was conducted. In drop-column feature importance analysis, as the name suggests, a column or feature is systematically removed from the dataset to gauge its impact on prediction accuracy. Consequently, a feature's importance value is determined as the disparity between the $R^2$ score of the baseline featurization and the model after excluding that feature. It is worth noting that a negative importance value indicates that the removal of that feature actually enhances the model's performance. The findings of the drop-column feature importance investigation are depicted in Figure \ref{fig:fig8}. Notably, the post-processing condition—such as as-built, heat-treated, heat isostatic presses, and stress-relieved—emerges as the most influential feature affecting the mechanical properties of additively manufactured parts. Another noteworthy observation is that orientation exhibits a minor effect on yield strength and ultimate tensile strength, while significantly impacting elongation at break, consistent with the findings of Hrabe et al \cite{HRABE2013271}. Furthermore, processing properties like beam power and layer thickness were deemed important features across all tasks, followed by process, machine type, specimen orientation, and material thermal properties.

\subsection{\normalfont  XGBoost feature importance}
\begin{figure}\centering
\includegraphics[scale=0.06]{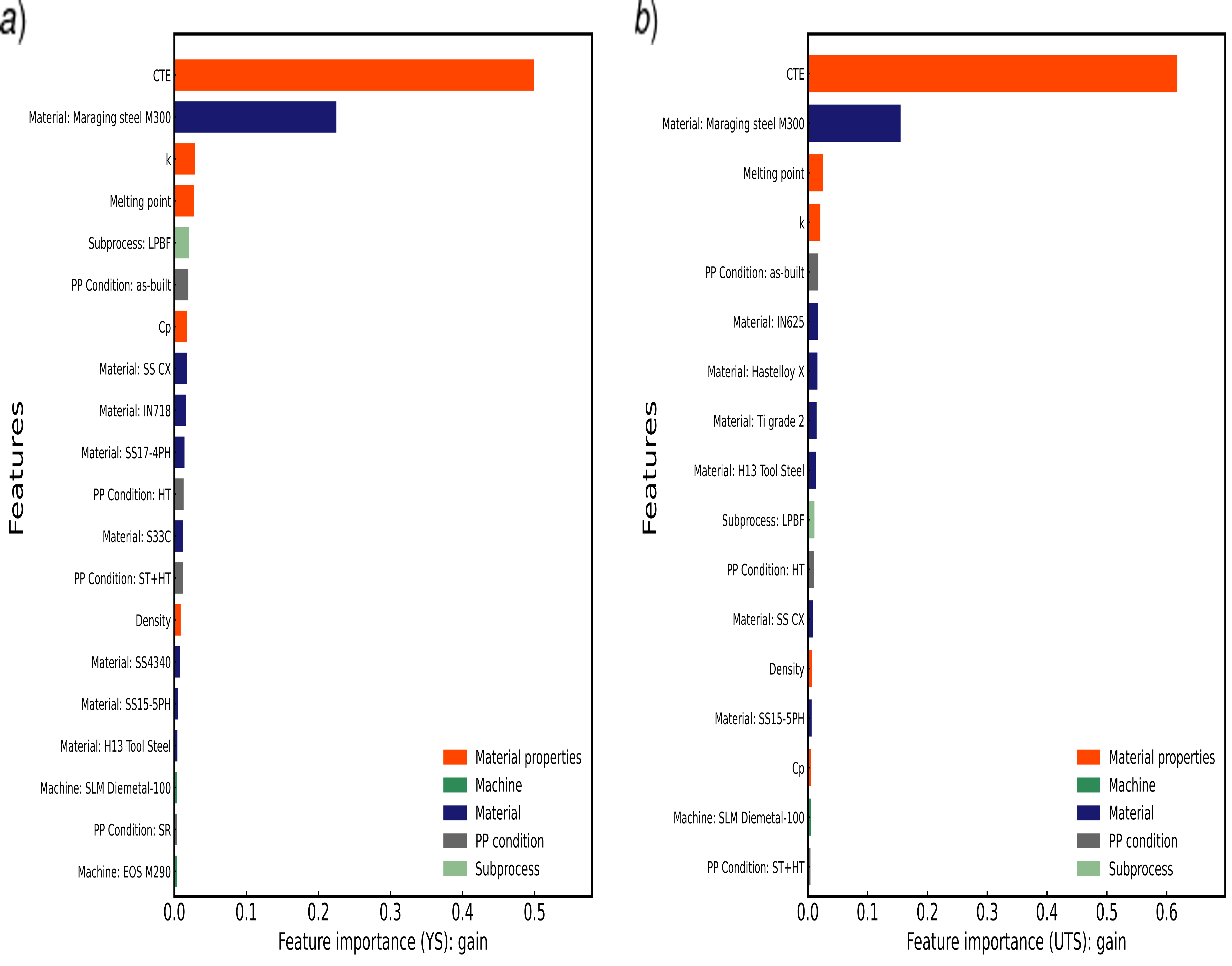}
\caption{Feature importance analysis using XGBoost for: a) Yield strength, and b) Ultimate tensile strength.}
\label{fig:fig9}
\end{figure}

An assessment of the relative contributions of each feature to the model prediction was conducted through XGBoost feature importance analysis. The importance type employed was "gain," wherein a higher value of this metric for a specific feature indicates its greater importance in generating predictions. Figure \ref{fig:fig9} provides a summary of the XGBoost feature importance results for yield strength and ultimate tensile strength. Given that tree-based models exhibited the most accurate prediction performances on our datasets, the XGBoost model was utilize for subsequent SHAP analysis in the following section.

\subsection{\normalfont Explainable AI}
In scientific applications, there is a particular interest in interpretability, which refers to the degree to which the rationale behind a model's decisions and predictions can be easily comprehended by humans. While machine learning models offer reliable predictions for a variety of tasks, they often lack interpretability and are commonly referred to as black-box models. To address this challenge, the Explainable AI technique known as Shapley Additive Explanations (SHAP), introduced by Lundberg et al., was employed\cite{NIPS2017_7062}, to elucidate the XGBoost prediction results in a manner that can be understood by humans. SHAP is a method that elucidates machine learning outputs by employing a game theoretic approach, wherein the features of a data point are treated as players in a coalition, facilitating interpretation.
\vspace{3mm}

SHAP, or Shapley Additive Explanations, assigns a Shapley value to each feature for a specific prediction, representing the average contribution of that feature value to the prediction across all possible feature combinations. In essence, the Shapley value of a feature quantifies the average change in the prediction when that feature is included within various feature combinations. Therefore, by utilizing TreeExplainer SHAP analysis \cite{lundberg2020local2global}, we were able to provide intuitive explanations for why a particular prediction was made, irrespective of the complexity of the ML model employed. It is important to note that Shapley values do not signify the difference in prediction when a feature is removed from the entire set of features. As such, the results from SHAP analysis differ from those obtained in the drop-column feature importance study conducted in the previous section. While drop-column feature importance is based on the decrease in model performance resulting from the removal of a feature, SHAP analysis focuses on the magnitude of feature attributions in influencing predictions.

\vspace{3mm}

Although feature importance plots offer valuable insights, they often lack detailed information beyond feature importance scores. To delve deeper into the behavior of individual features across observations, various SHAP plots can be utilized. While tree-based methods like XGBoost provide an overall understanding of how input features impact model predictions, they fall short in offering local explanations for the influence of input features on individual predictions \cite{lundberg2020local2global}. Therefore, employing SHAP plots allows us to gain comprehensive insights into the behavior of machine learning models, both globally and locally.

\vspace{3mm}

\begin{figure}\centering
\includegraphics[scale=0.13]{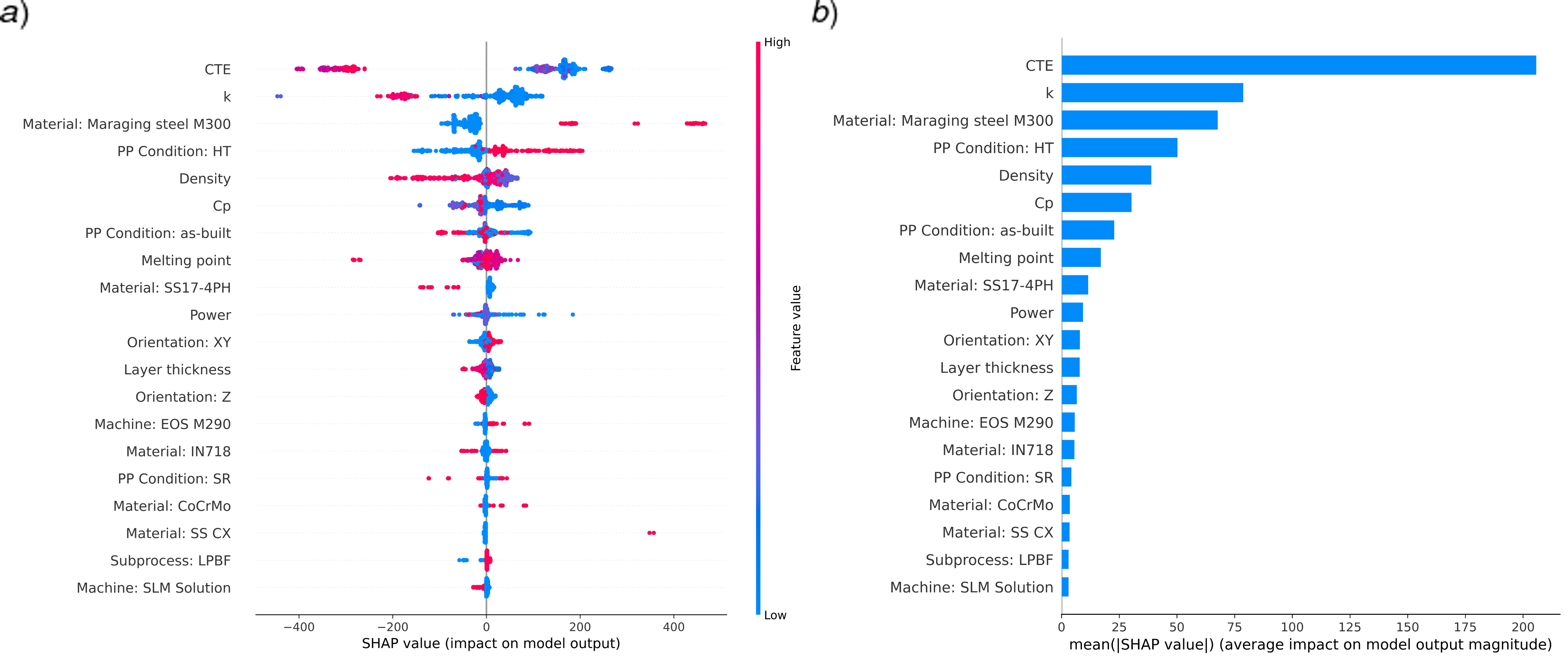}
\caption{Summary plot of SHAP analysis for the XGBoost model predicting yield strength. a) A swarm plot illustrating the influence of each feature on predicted yield strength; the color of dots indicates the relative value of the feature within the dataset (ranging from low in pink to high in blue), while the position of dots on the x-axis indicates whether the impact of that feature value positively or negatively contributed to yield strength prediction. b) Mean SHAP plot displaying the average of absolute SHAP values across all data points.}
\label{fig:fig10}
\end{figure}

Fig. \ref{fig:fig10}a depicts a SHAP summary plot that integrates feature importance with feature effects, offering a global perspective of SHAP values regarding each feature's contribution to yield strength prediction compared to the average model prediction. On the right-hand side of the plot, the y-axis displays corresponding feature values, ranging from low to high. Each dot represents the Shapley value for a feature in a single observation from our dataset.  Features are arranged based on their importance, allowing for the inference of the relationship between feature values and their impact on predictions. For example, for features like CTE and k, an increase in feature value corresponds to a decrease in SHAP value, indicating that higher values for CTE and k lead to lower predicted yield strength. Additionally, since material, machine, and post-processing conditions have been one-hot encoded, red points represent a value of 1, while blue points represent a value of 0 in the dataset. Consequently, if the material under consideration for yield strength prediction is Maraging Steel M300, or if the additively manufactured part has undergone heat treatment after fabrication, the predicted yield strength is significantly higher than the average model prediction. Moreover, relationships between features can be inferred by examining SHAP plots; for instance, CTE and k exhibit an inverse relationship with heat treatment.

\vspace{3mm}

In addition, the mean SHAP plot (Fig. \ref{fig:fig10}b)  illustrates the average of absolute SHAP values across all data points, ensuring that positive and negative values do not cancel each other out. Each bar corresponds to a feature in the dataset, sorted based on its impact on the model's prediction. As such, this plot can be interpreted as a SHAP feature importance plot. However, it does not provide any insight into the specific nature of the relationship between features and yield strength.

\vspace{3mm}

Moreover, the SHAP waterfall plot provides a visual representation of SHAP values for features in an individual prediction pertaining to a specific row in the dataset. It illustrates the extent to which each feature has influenced the model's prediction compared to the mean prediction, thereby either increasing or decreasing the predicted yield strength for that particular data point (refer to Table \ref{tab:datapoint}). As depicted in Fig. \ref{fig:fig11}a, the base value at the bottom of the waterfall plot begins at E[f(x)] = 789.328 MPa, representing the average predicted yield strength across all datasets. Each row demonstrates how the positive (in red) or negative (in blue) contribution of each feature has shifted the expected model output—the base value—to the model-predicted yield strength for that data point, which in this instance is f(x) = 923.825 MPa.

\vspace{3mm}

For example, considering the 300th data point in the yield strength dataset as presented in Table \ref{tab:datapoint}, small values for CTE and k contribute to an increase in the predicted target, thereby boosting the predicted yield strength by 176.36 $MPa$ and 61.23 $MPa$, respectively, as observed in Fig. \ref{fig:fig11}a and detailed in Table \ref{tab:datapoint}. Additionally, since the material for this data point is Ti6Al4V ELI, the value for Maraging Steel is zero, resulting in a decrease in the predicted value by 70.64 MPa. Furthermore, the post-processing method for this observation— as-built—has further decreased the prediction, aligning with the findings in Figure \ref{fig:fig10}a. It is important to note that for each data point in the dataset, a unique SHAP waterfall plot will be generated.

\vspace{3mm}

\begin{table}[ht]
\begin{center}
\caption{Data point No. 300 details in the yield strength dataset.}\vspace{1mm}
\label{tab:datapoint}
\scalebox{0.6}{
\begin{tabular}{c c c  c c c c }
\toprule [1pt]
Material & Process & Subprocess & Machine type &  Orientation & PP Condition  &  \\
\midrule [2pt]
Ti6Al4V ELI & PBF & L-PBF & EOS M4OO SF & XY & as-built \\
\toprule [0.5pt]
\toprule [0.5pt]
Power& Layer thickness & Density & Specific heat & Thermal conductivity & CTE & Melting point  \\
\midrule [2pt]
1000 &	30	& 4.41&		526 & 7 &		8.5 & 1650\\
\bottomrule
\end{tabular} 
}
\end{center}
\end{table}

A SHAP force plot offers an alternative visualization of SHAP values, conveying nearly identical information as a waterfall plot. In Fig. \ref{fig:fig11}b, once again focusing on the same data instance (as detailed in Table \ref{tab:datapoint}), it is observed that it initiates from the same base value of 789.328 MPa. It demonstrates how each feature contributes to either an increase or decrease in the predicted value, ultimately resulting in the final prediction of 923.825 MPa. By constructing force plots for every data point in the dataset, rotating them 90 degrees, and then aligning them horizontally, a multi-prediction force plot can be generated, illustrating explanations for the entire dataset, as depicted in Fig. \ref{fig:fig11}c.

\begin{figure}\centering
\includegraphics[scale=0.065]{Figures/figure_11.jpg}
\caption{a) SHAP waterfall plot illustrating the contribution of each feature to the model's prediction for a particular data point in the yield strength dataset using the XGBoost model. b) SHAP force plot for the same data sample showcasing the breakdown of the predicted yield strength value (f(x)) into its base value and the contributions from individual features. c) Multi-prediction force plot illustrating the collective behavior of the entire yield strength dataset.}
\label{fig:fig11}
\end{figure}
\vspace{3mm}

\begin{figure}\centering
\includegraphics[scale=0.11]{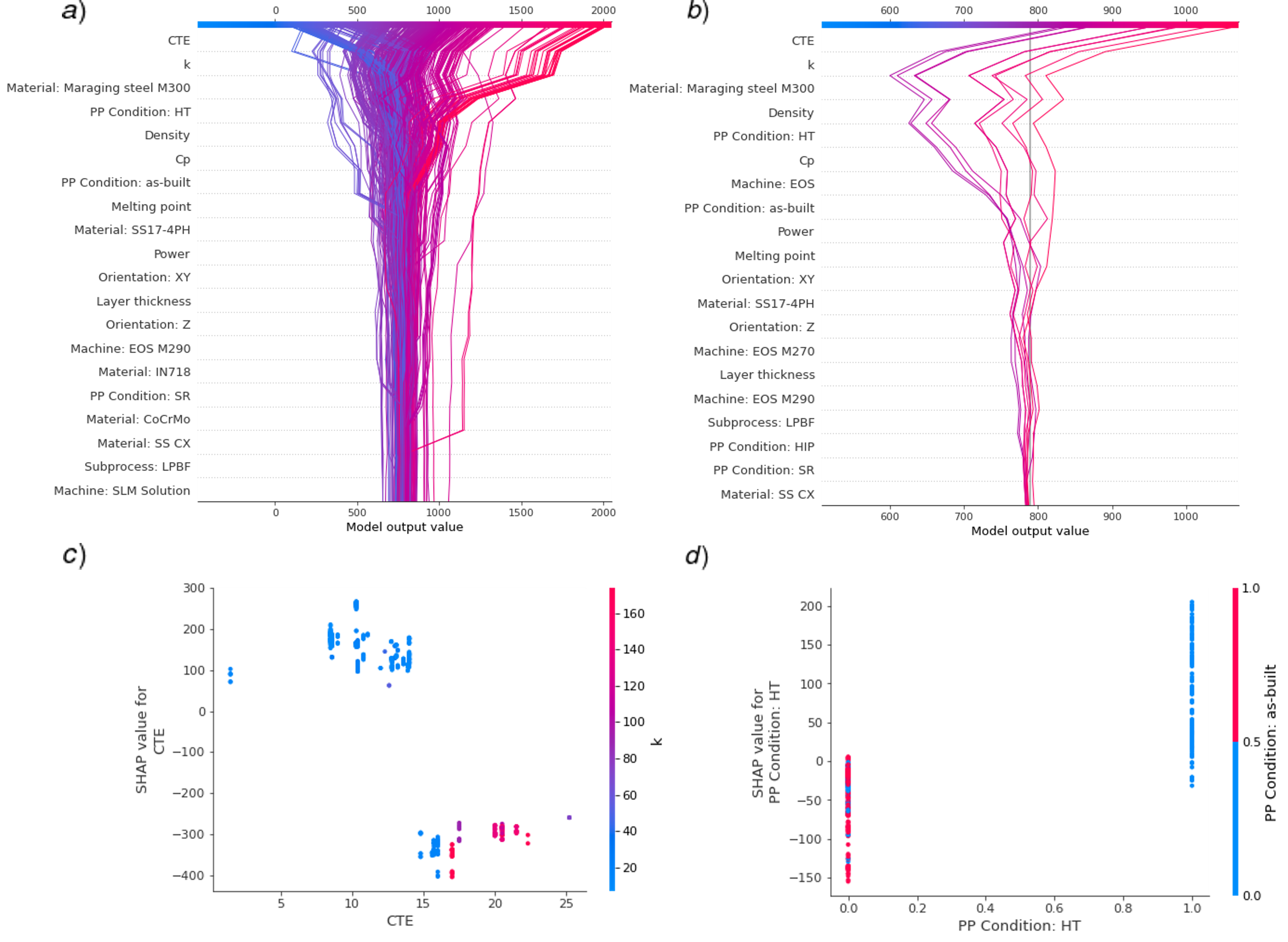}
\caption{a) SHAP decision plot illustrating the decision path followed for each predicted yield strength across the entire benchmark dataset. b) SHAP decision plot displaying the decision path taken for each predicted yield strength for a subset of 10 data samples. c) SHAP dependence plot showcasing the effects of CTE on the predicted yield strength. d) SHAP dependence plot demonstrating the impact of heat treatment as a post-processing method on the predicted yield strength.}
\label{fig:fig12}
\end{figure}

To gain insights into how the ML model generates predictions overall, the SHAP decision plot was examined (Fig. \ref{fig:fig12}a,b). In these plots, the straight vertical line positioned at the bottom of the figure represents the model's base value. Each line, originating from this base value, illustrates how the SHAP value of each feature influences the prediction, ultimately leading to the model's final predicted yield strength at the top of the plot, representing one observation. It is worth noting that a decision plot for a single row of the dataset yields results similar to those obtained from waterfall and force plots. However, when exploring the collective impact of global SHAP values on predictions, while a multi-prediction force plot can be utilized, interpreting a decision plot is often more straightforward, especially for large datasets (Fig. \ref{fig:fig12}a).

\vspace{3mm}

Fig. \ref{fig:fig12}b presents the decision plot for 10 observations. Here, the lines at the top of the plot exhibit a zig-zag pattern, indicating that features such as CTE, k, and density contribute positively to the prediction (i.e., positive SHAP value), while the Material and post-processing HT exert a negative influence on the prediction (i.e., negative SHAP value). Essentially, these features demonstrate opposite effects on the prediction. Thus, these findings align with the insights derived from Fig. \ref{fig:fig10}a.

\vspace{3mm}

Additionally, the SHAP dependence plot, which illustrates the effect of a specific feature on the model's prediction, was analyzed. Similar to the SHAP summary plot, in the dependence scatter plot, each dot corresponds to a single row from the dataset. The x-axis represents the value of the specified feature across the entire dataset, while the y-axis denotes the SHAP value for that feature, indicating how the feature's value influences the model's prediction. The vertical coloring pattern also signifies the interaction effects of a second feature with the current feature.

\vspace{3mm}

Figures \ref{fig:fig12}c and \ref{fig:fig12}d demonstrate the impact of changes in CTE and heat treatment, respectively, on the predicted yield strength. It is evident that values larger than $15 \cdot 10^-6/K$ for CTE result in smaller yield strength, while materials with lower CTE values tend to have greater predicted yield strength. Moreover, for materials in our dataset with larger CTE values, higher thermal conductivity can be expected as well (Fig. \ref{fig:fig12}c).

\vspace{3mm}

Furthermore, Figure \ref{fig:fig12}d illustrates that heat treatment yields a positive SHAP value, indicating that implementing heat treatment as a post-processing method will increase the predicted yield strength according to the ML model. Conversely, when heat treatment is absent (i.e., zero values), it can be observed that a portion of the observations belongs to as-built materials, which do not undergo any post-processing methods, within the benchmark dataset.

\subsection{\normalfont Impact of Dataset Size on Accuracy}
\begin{figure}\centering
\includegraphics[scale=0.06]{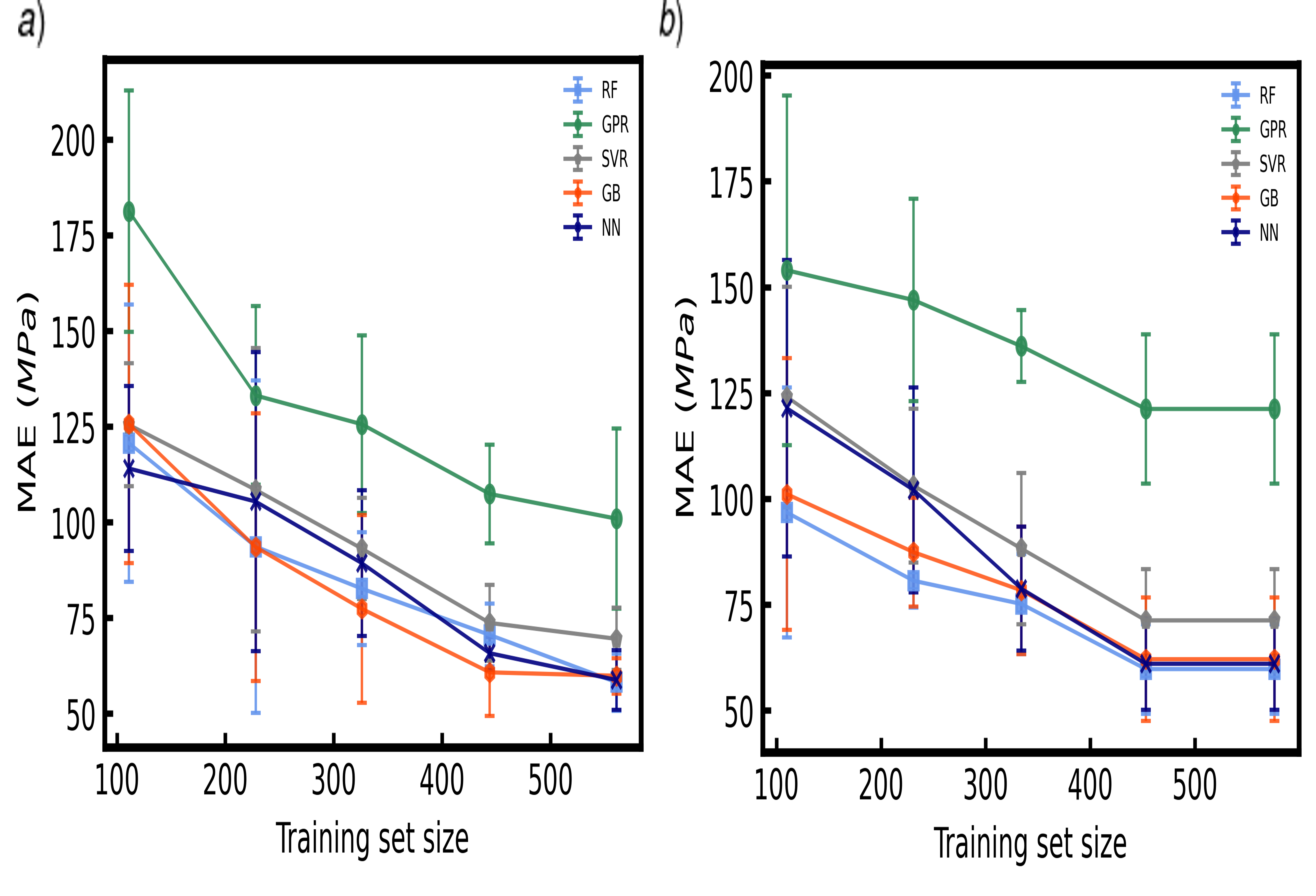}
\caption{Out-of-sample performances with different training set sizes with baseline featurization on a) yield strength b) ultimate tensile strength. (Each data point is the average of 5 independent runs, with standard deviations shown as error bars.)}
\label{fig:fig13}
\end{figure}

An examination was conducted to assess how varying the size of the training dataset influences the performance of regression tasks. To achieve this, random selections of 20\%, 40\%, 60\%, 80\%, and 100\% of samples from the training partition of the dataset were made, followed by the implementation of 5-fold cross-validation for various ML models. Figure \ref{fig:fig13} presents the mean and standard deviations of the test Mean Absolute Error (MAE) as a function of training sample size for both yield strength and ultimate tensile strength with baseline featurization. It is evident from the plot that as the size of the training set increases, there is a corresponding decrease in MAE, leading to an improvement in prediction accuracy.

\subsection{\normalfont Model Identification}

While the previous sections attempted to elucidate the prediction of yield strength by ML models, this section introduces a data-driven model identification approach aimed at extracting explicit relationships among processing parameters, material properties, and mechanical properties within the dataset. Unlike ML models, this method offers greater interpretability by providing mathematical equations.

\vspace{3mm} 

To establish mathematical and quantitative models from observed systems, various equation discovery methods are employed. For instance, symbolic regression techniques \cite{ai-feynman} have been utilized to formulate mathematical terms within equations. However, the complexity of managing numerous determinant parameters in MAM processes renders the application of current symbolic regression methods impractical, as they are primarily designed for cases involving a single covariate. Furthermore, the intricate nature of MAM processes and the diverse mathematical forms of equations pose challenges in employing parsimonious models within a sparse regression framework \cite{Sindy}. Therefore, a customized identification framework has been proposed to address these issues and handle multiple parameters under certain prior assumptions.

\vspace{3mm}  

In line with the regression tasks, our aim is to develop a predictive model for mechanical properties such as yield strength, ultimate tensile strength, and elastic modulus within the L-PBF dataset. While numerous parameters may influence mechanical properties, the investigation is focused on a specific set of processing parameters and material properties. The dimensions of these parameters are outlined in Table \ref{table:dim}. Considering the significance of post-processing conditions in predicting mechanical properties, distinct models for different post-processing conditions, namely, as-built and heat-treated conditions, are identified.

\begin{table}[ht]
\centering
\caption{List of parameters and their dimensions employed in data-driven model identification}
\begin{tabular}{c|c|c c c c}
\toprule [1pt]
Parameters & Unit (SI) &$kg$ & $m$ & $s$ & $K$ \\
\midrule [2pt]
YS, UTS, E & Pa ($\frac{kg}{ms^2})$ &1 & -1 & -2 & 0 \\
\midrule [1pt]
Beam power (P) & W & 1 & 2 & -3 & 0 \\[3pt]
Scanning speed (V) & $\frac{m}{s}$ & 0 & 1 & -1 & 0  \\[3pt]
Layer thickness (L) & $m$ & 0 & 1 & 0 & 0  \\[3pt]
Beam diameter (B) &$m$  & 0 & 1 & 0 & 0  \\[3pt]
Density ($\rho$) & $\frac{kg}{m^3}$ & 1 & -3 & 0 & 0 \\[3pt]
Specific heat ($C_p$) & $\frac{m^2}{s^2 K}$ & 0 & 2 & -2 & -1\\[3pt]
Coefficient of thermal expansion ($CTE$) & $\frac{1}{K}$ & 0 & 0 & 0 & -1\\
Thermal conductivity ($k$) & $\frac{kg m}{s^3 K}$ & 1 & 1 & -3 & -1\\[3pt]
Melting temperature ($T_m$) & $K$& 0 & 0 & 0 & 1\\
\bottomrule
\end{tabular}
\label{table:dim}
\end{table}

\vspace{3mm} 

Dimensional analysis is crucial in physical equations due to the necessity for mathematical equations describing physical quantities to maintain dimensional consistency. Given the presence of parameters with various dimensions in our equation, simple linear regression fails to uphold dimensional consistency on both sides of the equation. Therefore, a nonlinear regression model is introduced, formulated as an optimization problem, with dimensional consistency serving as its constraints. This nonlinear form involves multiplying a constant variable, represented by a multiplier ($w_0$), with the exponent variables for each parameter (Eq. \ref{eq:nonlinear-form}). Four constraints are devised to ensure the satisfaction of dimension consistency for each basic unit of mass, length, time, and temperature, as outlined in Table \ref{table:dim}. Subsequently, our objective is to minimize the $L_2$ norm of the equation, serving as the objective function, subject to the compacted form of these constraints (Eq. \ref{eq:optimization}).

\vspace{3mm} 

\begin{equation}
\begin{aligned}
 y = & w_0. P^{w_1} V^{w_2} L^{w_3} B^{w_4} \rho^{w_5} C_p^{w_6} CTE^{w_7} k^{w_8} (T_m - T_0)^{w_9} \\
\textrm{where}: \quad & 1) w_1 +w_5 +w_8 = 1 \\
& 2) 2w_1 + w_2 + w_3 + w_4 - 3w_5 + 2w_6 + w_8 = -1 \\
& 3) -3w_1 - w_2 - 2w_6 - 3w_8 = -2 \\
& 4) w_6 - w_7 - w_8 + w_9 = 0
\end{aligned}
\label{eq:nonlinear-form}
\end{equation}

\begin{equation}
\begin{aligned}
\min_{W} \quad & \| y -  w_0 P^{w_1} V^{w_2} L^{w_3} B^{w_4} \rho^{w_5} C_p^{w_6} CTE^{w_7} k^{w_8} (T_m - T_0)^{w_9} \|_2^2 \\
\textrm{s.t.}: \quad & (w_1 +w_5 +w_8 -1)^2 + (2w_1 + w_2 + w_3 + w_4 - 3w_5 + 2w_6 + w_8 +1)^2 +\\
& (-3w_1 - w_2 - 2w_6 - 3w_8 +2)^2 + (w_6 - w_7 - w_8 + w_9)^2 = 0 \\
\end{aligned}
\label{eq:optimization}
\end{equation}

\vspace{3mm} 

The SciPy \cite{2020SciPy-NMeth} constrained optimization package, utilizing trust-region methods \cite{trust-region} was employed, to address this optimization problem. Trust-region methods are iterative optimization algorithms that aim to minimize a function within a region where a quadratic model provides a good approximation to the objective function. These methods iteratively construct and update a quadratic model of the objective function and restrict the optimization to a trust region around the current iterate, ensuring that the model remains accurate within this region. By iteratively refining the trust region and updating the quadratic model, trust-region methods efficiently navigate the search space to find local minima. It is important to note that solutions are not guaranteed to reach a global minimum, hence, a sample solution for each task is provided in Table \ref{table:eq-results}.

\vspace{3mm}

To facilitate a comparison of prediction performance among different models, the $R^2$ scores of the identified equations alongside those of the best ML model is reported in Table \ref{table:comparison}. While ML models may lack interpretability, they typically offer superior predictive performance. Thus, despite the interpretability challenge, they remain a benchmark for comparison.

\begin{table}[ht]
\centering
\caption{Identified equations for modeling mechanical properties and their $R^2$.}
\scalebox{0.68}{
\begin{tabular}{c|c|c|c}
\toprule [1pt]
Task  & Post-processing & Equation &$R^2$ \\
(label) & condition & & \\
\midrule [2pt]
YS & As-built &  $YS= 0.83 \times 10^6 \times P^{0.07} V^{-0.05} L^{-0.18} B^{-0.08}\rho^{0.82} C_p^{0.75} CTE^{-0.94} k^{0.11} (T_m - T_0)^{-0.29} $ & 0.91  \\

& HT &$YS=  7.18 \times P^{1.43} V^{-0.66} L^{-1.1} B^{-0.67}\rho^{0.65} C_p^{0.81} CTE^{-1.99} k^{-0.10} (T_m - T_0)^{-1.08} $& 0.91\\
\midrule [0.5pt]

UTS & As-built & $UTS =  0.65 \times 10^6 \times P^{0.04} V^{-0.05} L^{-0.19} B^{-0.03}\rho^{0.81} C_p^{0.74} CTE^{-0.78} k^{-0.18} (T_m - T_0)^{0.15}$ & 0.89   \\

& HT & $UTS =  1.18 \times 10^6 \times P^{0.49} V^{-0.13} L^{-0.31} B^{-0.07}\rho^{1.11} C_p^{1.23} CTE^{-1.99} k^{-0.15} (T_m - T_0)^{-0.6} $& 0.853 \\
\midrule [1pt]

E & As-built & $E =  1.95 \times 10^6 \times P^{0.55} V^{-0.20} L^{-0.44} B^{-0.23}\rho^{0.88} C_p^{0.92} CTE^{-1.26} k^{0.09} (T_m - T_0)^{-0.43}$ & 0.848\\

& HT &$E =  0.97 \times 10^6 \times P^{0.37} V^{-0.21} L^{0.06} B^{-0.78}\rho^{0.65} C_p^{0.58} CTE^{-0.59} k^{0.01} (T_m - T_0)^{-0.02}$ & 0.994
\\
\bottomrule
\end{tabular}
}
\label{table:eq-results}
\end{table}

\begin{table}[ht]
\centering
\caption{$R^2$ accuracy comparison for the identified equations and ML models for yield strength, ultimate tensile strength, and elastic modulus.}
\scalebox{0.9}{
\begin{tabular}{c|c|c}
\toprule [1pt]
Task (label) & Identified equation $R^2$ & ML $R^2$\\
\midrule [2pt]
Yield strength &  0.91 & 0.958 \\

Ultimate tensile strength & 0.853 & 0.9694  \\

Elastic modulus & 0.848 &  0.9427  \\
\bottomrule
\end{tabular}
}
\label{table:comparison}
\end{table}

\vspace{3mm}

\section{Conclusion}

This study presents an extensive machine learning benchmark for predicting the mechanical properties of additively manufactured parts. A comprehensive experimental dataset encompassing various MAM processes, materials, and machines was assembled, specifically focusing on their effects on mechanical properties. Several featurization techniques were explored to enhance the performance of multiple ML models, while also examining evaluation metrics and hyperparameter optimization methods.

\vspace{3mm}

Among the ML algorithms tested, including Random Forest, Gradient Boosting, and Neural Networks, it was found that these models consistently outperformed others. Consequently, an accurate benchmark ML model capable of predicting mechanical properties in MAM processes was successfully developed, offering generalizability across different materials, processes, and machines.

\vspace{3mm}

Additionally, an effort was made to enhance interpretability by investigating the Explainable AI method, specifically SHAP analysis, to elucidate the prediction of yield strength by an ML model. Furthermore, a data-driven model identification method was devised to establish explicit models for mechanical properties based on dataset processing parameters and material properties, aiming for greater interpretability compared to the employed ML models. By providing a standardized platform for comparison and evaluation, we anticipate that our benchmark, the MechProNet, will facilitate the optimization of additive manufacturing processes and serve as a valuable resource for the metal additive manufacturing machine learning community.

\section*{Declaration of competing interest}
The authors declare that they have no known competing financial interests or personal relationships that could have appeared to influence the work reported in this paper.

\section*{Acknowledgements}
Research was sponsored by the Army Research Laboratory and was accomplished under Cooperative Agreement Number W911NF-20-2-0175. The views and conclusions contained in this document are those of the authors and should not be interpreted as representing the official policies, either expressed or implied, of the Army Research Laboratory or the U.S. Government. The U.S. Government is authorized to reproduce and distribute reprints for Government purposes notwithstanding any copyright notation herein. AM acknowledges the startup funding provided by the Mechanical, Materials, and Aerospace Engineering Department at the Illinois Institute of Technology in Chicago, Illinois. AM acknowledges partial support from the National Science Foundation under grant number DMR-2050916 and CMMI-2339857.

\setcounter{table}{0}

\appendix

\section{Supplementary data}
\label{sec:sample:appendix}
The code and data associated with this article will be available online upon publication.

\begin{table}[hbt!]
\begin{center}
\caption{Chemical composition (wt\%) of alloys studied in our benchmark.}\vspace{1mm}
\label{sec:sample:appendix:tab:tableA1}
\scalebox{0.5}{
\begin{tabular}{c c c c c c c c c c c c c c c c c c c c c}
\toprule [1pt]
{alloys} & Fe & Cr & Ni & Mo & Al & Mn & Si & Mg & Nb & Ti & V & Co & Zr & Cu & W & N & C & Sn & Zn & Sc \\ [1ex]
\midrule [2pt]
Ti6Al4V & 0&	0&	0&	0&	6&	0&	0&	0&	0&	90&	4&	0&	0&	0&	0&	0&	0&	0&	0&	0\\
IN718 &15&	19	&52.5&	3&	0&	0&	0&	0&	5	&1&	0&	0&	0&	0&	0&	0&	0&	0&	0&	0\\
SS316L& 65&	17&	12&	2.5&	0&	0&	0&	0&	0&	0&	0&	0&	0&	0&	0&	0&	0&	0&	0&	0\\
AlSi10Mg& 0&	0&	0&	0	&89	&0&	10&	0.35&	0&	0&	0&	0&	0&	0&	0&	0&	0&	0&	0&	0\\
IN625 &5	&22&	60&	9&	0	&0	&0&	0&	0	&0&	0&	0&	0&	0&	0	&0&	0&	0	&0&	0\\
Maraging steel M300&0&	0&	18&	5&	0	&0	&0&	0&	0&	0.6	&0&	9&	0&	0&	0&	0&	0.02&	0&	0&	0\\
Ti6Al4V ELI &0&	0&	0&	0&	6&	0	&0	&0&	0&	90	&4&	0&	0&	0&	0&	0&	0&	0&	0&	0\\
SS17-4PH& 74.5&	16&	4&	0&	0&	1	&0&	0&	0.3	&0&	0&	0&	0&	4&	0&	0&	0&	0&	0&	0\\
CoCrMo & 0.5	&28.5&	0.4&	6	&0.05&	0&	0.7	&0.7&	0&	0.05&	0	&62.6&	0&	0&	0.1&	0&	0.35&	0&	0&	0\\
H13 Tool Steel &91&	5.1&	0&	1.4	&0&	0.4	&1&	0&	0&	0&	1&	0&	0&	0&	0&	0&	0.37&	0&	0&	0\\
AlSi7Mg0.6 &0&	0&	0&	0	&93.06	&0&	6.5	&0.4&	0	&0.04&	0&	0&	0&	0&	0&	0&	0&	0&	0	&0\\
SS15-5PH  &74.5&	15&	4.5	&0&	0&	1&	1&	0&	0.3&	0&	0&	0&	0	&3.5&	0&	0&	0&	0&	0&	0\\
CuCrZr &0&	1&	0&	0&	0&	0&	0&	0&	0&	0&	0	&0&	0.07&	98.7&	0&	0&	0&	0&	0&	0\\
Al12Si &0&	0	&0&	0&	12&	0&	88&	0&	0&	0&	0&	0&	0&	0&	0&	0&	0&	0&	0&	0\\
Copper &0.04	&0&	0.015&	0&	0&0&	0&	0&	0&	0&	0&	0&	0&	99.9&	0&	0&	0.02&	0&	0&	0\\
Hastelloy X &18.5&	21.75&	0	&9	&0.5&	1&	1&	0&	0&	0.15&	0&	1.5	&0&	0.5&	0.6&	0&	0&	0&	0&	0\\
CoCr &0&	28.5	&0&	6	&0&	0&	1&	0&	0&	0&	0&	61&	0&	0&	0&	0&	0&	0&	0	&0\\
Hastelloy C22 &4&	21	&0&	13.5&	55&	0.5	&0.08&	0&	0&	0&	0.35&	2.5	&0&	0&	3&	0&	0.015&	0&	0&	0\\
SS254& 56.32&	19.5&	17.5&	6&	0&	0&	0&	0&	0&	0&	0&	0&	0	&0.5	&0&	0.18&	0&	0&	0&	0\\
CuNi2SiCr& 0.15	&0.35&	2.5&	0&	0&	0.1&	0.65&	0&	0&	0	&0&0&	0&	96.15&	0&	0&	0&	0&	0&	0\\
D2 Tool Steel &83.8&	12	&0.45&	1&	0&	0.5	&0.3&	0	&0&	0&	0.9	&0&	0&	0&	0&	0&	1.5&	0&	0&	0\\
SS4340 &95.5&	0.8	&1.8&	0.25&	0&	0.7&	0.22&	0&	0&	0&	0&	0&	0&	0&	0&	0&	0.4	&0&	0&	0\\
Invar36& 62.7&	0.5	&36&	0&	0&	0.5&	0.5	&0&	0&	0&	0&	0&	0&	0&	0&	0&	0	&0&	0&	0\\
SS304L &68&	19	&10&	0&	0&	2&	0.75&	0&	0&	0&	0&	0&	0&	0&	0&	0.1	&0.03&	0&	0&	0\\
Ti grade 2 &0.3&	0&	0&	0&	0&	0&	0&	0&	0&	99&	0&	0&	0&	0&	0&	0	&0	&0&	0&	0\\
K500& 0&	0&	70.35&	0&	2.3	&0&	0&	0&	0&	0.35&	0&	0&	0&	27&	0&	0&	0&	0&	0&	0\\
M2 Tool Steel &79&	4.1	&0.2&	5&	0&	0.3&	0.33&	0&	0&	0&	2&	0&	0&	0&6.1&	0&	0.9	&0&	0&	0\\
Ti grade 1& 0&	0&	0&	0&	0&	0&	0&	0&	0&	100&	0&	0&	0&	0&	0&	0&	0&	0&	0&	0\\
CoCr28Mo6 &0.75&	28.5&	0.5	&6&	0.1	&1&	1&	0&	0&	0&	0&	62	&0&	0	&0&	0&	0&	0&	0&	0\\
IN690& 9&	29&	60.5&	0&	0&	0.5	&0.5&	0&	0&	0&	0&	0&	0&	0.5&	0&	0&	0.05&	0&	0&	0\\
CuCP &0	&0&	0&	0&	0&	0&	0&	0&	0&	0&	0&	100	&0&	0&	0&	0	&0&	0&	0&	0\\
20MnCr5 &97.73&	1&	0&	0&	0&	1.1&	0&	0&	0&	0&	0&	0&	0&	0&	0&	0&	0.17&	0&	0&	0\\
TiAlNbMo &0	&0	&0&	2.26&	28.9&	0&	0&	0&	9.68&	59.16&	0&	0&0&	0&	0&	0&	0&	0&	0&	0\\
SS304& 68.5&	19	&9.25&	0&	0&	2&	0.75&	0&	0&	0&	0&	0&	0&	0&	0&	0.1&	0.08&	0&	0&	0\\
TiAl& 0&	2.2&	0&	0	&32&	0&	0&	0&	4.5	&61.3&	0&	0&	0&	0&	0&	0&	0	&0&	0&	0\\
HY100& 93.5&	1.7	&3.25&	0.57&	0&	0.25&	0.26&	0&	0&	0&	0&	0&	0&	0&	0&	0&	0.17&	0&	0&	0\\
AlSi12& 0&	0&	0&	0&	88&	0&	12&	0&	0&	0&	0&	0&	0&	0&	0&	0&	0&	0&	0&	0\\
CoCrF75& 0&	27&	0&	5&	0&	0&	0&	0&	0&	0&	0&	68&	0&	0&	0&	0&	0&	0&	0&	0\\
AZ91D& 0&	0&	0&	0&	8.95&	0.19&	0&90.42&	0	&0&	0&	0&	0&	0&	0&	0&	0&	0&	0.44&	0\\
IN939& 0&	22.5&	48.5&	0&	1.9&	0&	0&	0&	2.4&	3.7&	0&	19&	0&	0&	2&	0&	0&	0&	0&	0\\
CoCrW &0&	28&	0&	0&	0&	0&1.5&	0&	0&	0&	0&	60.5&	0&	0&	9&	0&	0&	0&	0	&0\\
420steel-6-bronze-4 &50	&7.8	&0.06	&0&	0	&0.9	&0.6&	0&	0&	0&	0&	0&	0&	37&	0&	0&	0.12&	4&	0&	0\\
SS420C&84&	13&	0.5&	0.5&	0&	1&	1&	0&	0&	0&	0&	0&	0&	0&	0&	0&	0.15&	0&	0&	0\\
SS4130&98&	0.95&	0&	0&	0	&0.5	&0.22&	0&	0&	0&	0&	0&	0&	0&	0&	0&	0.3&	0&	0&	0\\
CuCrZrTi &0	&0.6&	0&	0&	0&	0&	0&	0&	0&	0.035&	0&	0&	0.035&	99&	0&	0&	0&	0&	0&	0\\
8620L Steel &97.77&	0.5	&0.55&	0.2	&0&	0.8	&0	&0&	0&	0&	0&	0&	0&	0&	0&	0&	0.175&	0&	0&	0\\
Al2319& 0.3	&0	&0&	0&	96&	0.3&	0.2	&0.02&	0&	0.15&	0.1&	0&	0.17&	6.3&	0&	0&	0&	0&	0.1	&0\\
Iron (0.004\%C) &99.99&	0&	0&	0&	0&	0&	0&	0	&0	&0&	0&	0&	0&	0&	0&	0&	0.004&	0&	0&	0\\
SS316 &65&	17&	12&	2.5	&0&	0&	0&	0&	0&	0&	0&	0&	0&	0&	0&	0&	0&	0&	0&	0\\
Iron (0.02\%C)& 99.98&	0&	0&	0&	0&	0&	0&	0&	0&	0&	0&	0&	0&	0&	0&	0&	0.02&	0&	0&	0\\
TC21 &0.05&	1.5&	0&	3&	6.5	&0&	0.09&	0&	1.9	&82&0&	0&	2.2	&0&	0&	0&	0&	2.2&	0&	0\\
Mar M 247& 0&	8.4	&60&	0.65&	5.4&	0&	0&	0&	0&	1&	0&	10&	0&	0&	9.8&	0&	0.13&	0&	0&	0\\
Scalmalloy &0.4	&0&	0&	0&	92&	0.5&	0.4&	4.65&	0&	0.15&	0.1&	0&	0&	0.1&	0&	0&	0&	0&	0.25&	0.74\\
W22Fe33Ni &2.2&	0&	3.3&	0&	0&	0&	0&	0&	0&	0&	0&	0&	0&	0&	94.5&	0&	0&	0&	0&	0\\
4140 Steel& 97&	1&	0&	0.25&	0&	0.8	&0.4&	0&	0&	0&	0&	0&	0&	0&	0&	0&	0.3&	0&	0&	0\\
Tungsten-Bronze &0&	0&	0&	0&	0&	0&	0&	0&	0&	0&	0&	0&	0&	43.5&	52.5&	0&	0&	4&	0&	0\\
S50C& 98.5&	0&	0&	0&	0&	0.79&	0.22&	0&	0&	0&	0&	0&	0&	0&	0&	0&	0.46&	0&	0&	0\\
S75C& 98.2&	0&	0&	0&	0&	0.77&	0.28&	0&	0&	0&	0&	0&	0&	0&	0&	0&	0.71&	0&	0&	0\\
A2 Tool Steel &91.4	&5.125&	0&	1.15&	0&	0.7&	0.3&	0&	0&	0&	0.33&	0&	0&	0&	0&	0&	1&	0&	0&	0\\
S105C &98&	0&	0&	0&	0&	0.74&	0.21&	0&	0&	0&	0&	0&	0&	0&	0&	0&	0.99&	0&	0&	0\\
SS316L-6-Bronze-4 &38&	11.5&	9&	1.58&	0&	1.2&	0.45&	0&	0&	0&	0&	0&	0&	37&	0&	0.06&	0.03&	4&	0&	0\\
TA15& 0.25&	0&	0&	1.25&	6.3	&0	&0.15&	0&	0&	88	&1.65&	0&	2&	0&	0&	0&	0&	0&	0&	0\\
SS CX &78.3&	11&	8.4	&1.1&	1.2	&0	&0	&0	&0&	0&	0&	0&	0&	0&	0&	0&	0	&0&	0&	0\\
7050 alloy& 0&	0&	0&	0&	89&	0&	0&	2.3	&0&	0&	0&	0&	0.12&	2.3&	0&	0&	0&	0&	6.2&	0\\
S33C& 98.5&	0&	0	&0	&0&	0.64&	0.19&	0&	0&	0&	0&	0&	0&	0&	0&	0&	0.31&	0&	0&	0\\
IN600& 8	&15.5&	74.4&	0&	0&	1&	0.5&	0&	0&	0&	0&	0&	0&	0.5&	0&	0&	0.1	&0&	0&	0\\
Nb &0&	0&	0&	0&	0&	0&	0&	0&	100	&0&	0&	0&	0&	0&	0&	0&	0&	0&	0&	0\\
\bottomrule 
\end{tabular} }
\end{center}
\end{table}

\begin{table}[hbt!]
\begin{center}
\caption{Thermal properties of alloys studied in our benchmark.}\vspace{1mm}
\label{sec:sample:appendix:tab:tableA2}
\scalebox{0.55}{
\begin{tabular}{c c c c c c}
\toprule [1pt]
{Alloys} & Density ($\frac{kg}{m^3}$) &  Thermal conductivity ($\frac{W}{m.K}$) & Melting temperature (C) & Coefficient of thermal expansion ($\frac {10^-6}{K}$) & Specific heat ($\frac{J}{kg.K}$) \\ [1ex]
\midrule [2pt]
Ti6Al4V & 4.43 &	7.1 &	1695 &	8.6	& 561.5\\
IN718 &  8.19  &	9 &	1298 &	14	& 435\\
SS316L &7.99	& 16.2	& 1385 &	16 &	500\\
AlSi10Mg& 2.68	&160 &	580	& 20.5 &	910\\
IN625& 8.44	& 9.2 &	1320 &	12.8 &	429\\
Maraging steel M300 & 8.1 &	14.2 &	1413 &	10.3 &	452\\
Ti6Al4V ELI & 4.42 &	7 &	1650 &	8.5	& 526 \\
SS17-4PH & 7.8	& 16 &	1420 &	10.4 &	460 \\
CoCrMo & 8.4 &	13 &	1390 &	14 &	450 \\
H13 Tool Steel &7.8	&28.6&	1427&	10.4&	460\\
AlSi7Mg0.6 &2.67&	140	&570	&21.5&	890\\
SS15-5PH& 7.8&	11&	1420&	10.8&	420\\
CuCrZr& 8.9	&320	&1085&	17&	380\\
Al12Si& 2.7&	175	&680&	22.3&	922\\
Copper &8.6	&350	&9.6	&17.01&	435\\
Hastelloy X &8.2&	9.1&	1355&	13.9&	486\\
CoCr& 8.3&	13&	1413.75	&12	&390\\
Hastelloy C22 &8.69	&8.9&	1399&	12.4&	414\\
SS254& 8.07&	14&	1355&	14.8&	500\\
CuNi2SiCr &8.84	&90	&1050&	17.5&	397\\
D2 Tool Steel& 7.7&	20&	1421	&10.3&	460\\
SS4340 &7.85&	44.5&	1427&	12.3&	475\\
Invar36 &8.1	&12.8	&1727&	1.5&	515\\
SS304L &8	&16.2&	1693&	17.3&	500\\
Ti grade 2 &4.5	&22&	1665&	8.6&	523\\
K500& 8.43&	17.5	&1350&	13.4&	418\\
M2 Tool Steel &8.055&	41.5&	4680&	10&	460\\
Ti grade 1 &4.51&	16	&1668&	7.17&	520\\
CoCr28Mo6& 8.47	&12.5&	1390&	13.2&	450\\
IN690  &8.19 &	13.5&	1360&	14.06&	450\\
CuCP& 8.9	&320&	1085&	17&	385\\
TiAlNbMo &4.112&	22	&1773&	12&	600\\
SS304 &8&	16.2&	1693&	17.3&	500\\
TiAl& 4.05&	15&	1460&	9&	600\\
HY100 &7.75&	34&	1520&	14&	460\\
AlSi12 &2.685&	140&	579&	20&	548.6\\
CoCrF75 &8.35&	14&	1390&	14	&390\\
AZ91D &1.81&	72&	533	&25.2&	1047\\
CoCrW &8.6	&15	&1370&	14.1&	390\\
SS420C &7.8	&24.9&	1480&	10.3&	460\\
SS4130 &7.85	&42.7&	1432&	14.6&	523\\
CuCrZrTi &8.9&	320	&1085&	17&	380\\
8620L Steel &7.85&	46.6	&1430&	6.6&475\\
Al2319& 2.84&	170	&600&	22.5&	864\\
Iron (0.004\%C) &7.86&	50	&1535&	7&	460\\
SS316 &7.99	&16.2&	1385&	16	&500\\
Iron (0.02\%C) &7.86&	50&	1535&	7&	460\\
Mar M 247 &7.8&	15&	1220&	12.8&	230\\
W22Fe33Ni& 17.6	&27	&3410&	4.5&	134\\
4140 Steel &7.85	&42.6&	1416&	12.2&	473\\
S50C &7.85&	21&	1380&	13.2&	486\\
S75C& 7.85&	25&	1400&	13&	480\\
A2 Tool Steel &7.86&	26&	1420&	10.6&	461\\
S105C &7.872&	53&	1430&	12.6&	481\\
SS CX &7.69&	16.2&	1450&	11.1&	502\\
7050 alloy &2.7	&180&	494	&23.5&	860\\
S33C& 7.85&	34	&1400&	13&475\\
IN600& 8.47	&14.8&	1380&	13.3&	444\\
Nb &8.4&	53.7&	2410&	7.1	&260\\
\bottomrule 
\end{tabular}
}
\end{center}
\end{table}

\begin{table}[ht!]
\begin{center}
\caption{Hyperparameters and their range studied in our benchmark ML models.}\vspace{2mm}
\label{sec:sample:appendix:tab:tableA3}
\scalebox{0.6}{
\begin{tabular}{c c c c c}
\toprule [1pt]
Regression ML Task & {Models} & Hyperparameters & Value &  Range studied   \\ [1ex]
\midrule [2pt]
Elongation at break & Random Forest & n\_estimators &	463 &1-500	\\ [2ex]
 & Support Vector Regressor & C & 982	&1-1000	\\ [0.8ex]
 &  & kernel & 'rbf'	&['linear', 'poly', 'rbf', 'sigmoid']	\\ [2ex]
 & Gradient boosting &n\_estimators & 462 &1-500	\\ [2ex]
 & Neural network & number of neurons &(128,32,64)	&[32, 64, 128, 256, 512]\\ [0.8ex]
 &  & alpha &0.04074280425105631	& 1e-7 -  1e-1\\ [2ex]
\midrule [0.5pt]
Hardness (Vickers) & Random Forest & n\_estimators &	492 &1-500	\\ [2ex]
 & Support Vector Regressor & C & 975	&1-1000	\\ [0.8ex]
 &  & kernel & 'rbf'	&['linear', 'poly', 'rbf', 'sigmoid']	\\ [2ex]
 & Gradient boosting &n\_estimators & 192 &1-500	\\ [2ex]
 & Neural network & number of neurons &(32,256,32)	&[32, 64, 128, 256, 512]\\ [0.8ex]
 &  & alpha &0.04925337472160725	& 1e-7 -  1e-1\\ [2ex]
\midrule [0.5pt]
Hardness (Rockwell) & Random Forest & n\_estimators &	297 &1-500	\\ [2ex]
 & Support Vector Regressor & C & 234	&1-1000	\\ [0.8ex]
 &  & kernel & 'poly'	&['linear', 'poly', 'rbf', 'sigmoid']	\\ [0.8ex]
   &  & degree & 3	&[2,3,4]	\\[2ex]
 & Gradient boosting &n\_estimators & 235 &1-500	\\ [2ex]
 & Neural network & number of neurons &(32,128,128)	&[32, 64, 128, 256, 512]\\ [0.8ex]
 &  & alpha &0.06155030321486836	& 1e-7 -  1e-1\\ [2ex]
\midrule [0.5pt]
Mean roughness depth & Random Forest & n\_estimators &	42 &1-500	\\ [2ex]
 & Support Vector Regressor & C & 42	&1-1000	\\ [0.8ex]
 &  & kernel & 'rbf'	&['linear', 'poly', 'rbf', 'sigmoid']	\\ [2ex]
 & Gradient boosting &n\_estimators & 42 &1-500	\\ [2ex]
 & Neural network & number of neurons &(512,32,128)	&[32, 64, 128, 256, 512]\\ [0.8ex]
 &  & alpha &0.06833260439770075	& 1e-7 -  1e-1\\ [2ex]
\bottomrule 
\end{tabular} 
}
\end{center}
\end{table}

\clearpage


\bibliographystyle{elsarticle-num} 
\bibliography{mybibliography}

\begin{thebibliography}{100}
\expandafter\ifx\csname url\endcsname\relax
  \def\url#1{\texttt{#1}}\fi
\expandafter\ifx\csname urlprefix\endcsname\relax\def\urlprefix{URL }\fi
\expandafter\ifx\csname href\endcsname\relax
  \def\href#1#2{#2} \def\path#1{#1}\fi

\bibitem{DDGU}
D.~D. Gu, W.~Meiners, K.~Wissenbach, R.~Poprawe, Laser additive manufacturing of metallic components: materials, processes and mechanisms, International Materials Reviews 57~(3) (2012) 133--164.

\bibitem{frazer}
W.~Frazier, Metal additive manufacturing: A review., J. of Materi Eng and Perform 23 (2014) 1917–1928.

\bibitem{GARDNER20232178}
L.~Gardner, Metal additive manufacturing in structural engineering – review, advances, opportunities and outlook, Structures 47 (2023) 2178--2193.

\bibitem{pasang2023additive}
T.~Pasang, A.~S. Budiman, J.~Wang, C.~Jiang, R.~Boyer, J.~Williams, W.~Z. Misiolek, Additive manufacturing of titanium alloys--enabling re-manufacturing of aerospace and biomedical components, Microelectronic Engineering 270 (2023) 111935.

\bibitem{aliyu2023laser}
A.~A.~A. Aliyu, C.~Panwisawas, J.~Shinjo, C.~Puncreobutr, R.~C. Reed, K.~Poungsiri, B.~Lohwongwatana, Laser-based additive manufacturing of bulk metallic glasses: recent advances and future perspectives for biomedical applications, Journal of Materials Research and Technology (2023).

\bibitem{zhao2023direct}
N.~Zhao, M.~Parthasarathy, S.~Patil, D.~Coates, K.~Myers, H.~Zhu, W.~Li, Direct additive manufacturing of metal parts for automotive applications, Journal of Manufacturing Systems 68 (2023) 368--375.

\bibitem{cai2023review}
Y.~Cai, J.~Xiong, H.~Chen, G.~Zhang, A review of in-situ monitoring and process control system in metal-based laser additive manufacturing, Journal of Manufacturing Systems 70 (2023) 309--326.

\bibitem{xie2023data}
Z.~Xie, F.~Chen, L.~Wang, W.~Ge, W.~Yan, Data-driven prediction of keyhole features in metal additive manufacturing based on physics-based simulation, Journal of Intelligent Manufacturing (2023) 1--14.

\bibitem{laleh2023heat}
M.~Laleh, E.~Sadeghi, R.~I. Revilla, Q.~Chao, N.~Haghdadi, A.~E. Hughes, W.~Xu, I.~De~Graeve, M.~Qian, I.~Gibson, et~al., Heat treatment for metal additive manufacturing, Progress in Materials Science 133 (2023) 101051.

\bibitem{sneha}
M.~Tang, P.~Pistorius, S.~e.~a. Narra, Rapid solidification: Selective laser melting of alsi10mg., JOM 68 (2016) 960–966.

\bibitem{ojo2023post}
O.~O. Ojo, E.~Taban, Post-processing treatments--microstructure--performance interrelationship of metal additive manufactured aerospace alloys: a review, Materials science and technology 39~(1) (2023) 1--41.

\bibitem{collins2016microstructural}
P.~Collins, D.~Brice, P.~Samimi, I.~Ghamarian, H.~Fraser, Microstructural control of additively manufactured metallic materials, Annual Review of Materials Research 46 (2016) 63--91.

\bibitem{brennan2021defects}
M.~Brennan, J.~Keist, T.~Palmer, Defects in metal additive manufacturing processes (2021).

\bibitem{brandl2011mechanical}
E.~Brandl, C.~Leyens, F.~Palm, Mechanical properties of additive manufactured ti-6al-4v using wire and powder based processes, in: IOP conference series: materials science and engineering, Vol.~26, IOP Publishing, 2011, p. 012004.

\bibitem{edwards2013electron}
P.~Edwards, A.~O'conner, M.~Ramulu, Electron beam additive manufacturing of titanium components: properties and performance, Journal of Manufacturing Science and Engineering 135~(6) (2013).

\bibitem{vrancken2012heat}
B.~Vrancken, L.~Thijs, J.-P. Kruth, J.~Van~Humbeeck, Heat treatment of ti6al4v produced by selective laser melting: Microstructure and mechanical properties, Journal of Alloys and Compounds 541 (2012) 177--185.

\bibitem{YADOLLAHI2015171}
A.~Yadollahi, N.~Shamsaei, S.~M. Thompson, D.~W. Seely, Effects of process time interval and heat treatment on the mechanical and microstructural properties of direct laser deposited 316l stainless steel, Materials Science and Engineering: A 644 (2015) 171--183.

\bibitem{yan2018data}
F.~Yan, Y.-C. Chan, A.~Saboo, J.~Shah, G.~B. Olson, W.~Chen, Data-driven prediction of mechanical properties in support of rapid certification of additively manufactured alloys, Computer Modeling in Engineering \& Sciences 117~(3) (2018) 343--366.

\bibitem{AKBARI2022102817}
P.~Akbari, F.~Ogoke, N.-Y. Kao, K.~Meidani, C.-Y. Yeh, W.~Lee, A.~B. Farimani, Meltpoolnet: Melt pool characteristic prediction in metal additive manufacturing using machine learning, Additive Manufacturing (2022) 102817.

\bibitem{ZHAN2021105941}
Z.~Zhan, H.~Li, Machine learning based fatigue life prediction with effects of additive manufacturing process parameters for printed ss 316l, International Journal of Fatigue 142 (2021) 105941.

\bibitem{xie}
X.~Xie, J.~Bennett, S.~Saha, Y.~Lu, J.~Cao, W.~K. Liu, Z.~Gan, Mechanistic data-driven prediction of as-built mechanical properties in metal additive manufacturing, npj Computational Materials 7~(1) (2021) 1--12.

\bibitem{carl}
C.~Herriott, A.~D. Spear, Predicting microstructure-dependent mechanical properties in additively manufactured metals with machine-and deep-learning methods, Computational Materials Science 175 (2020) 109599.

\bibitem{xia}
C.~Xia, Z.~Pan, J.~Polden, H.~Li, Y.~Xu, S.~Chen, Modelling and prediction of surface roughness in wire arc additive manufacturing using machine learning, Journal of Intelligent Manufacturing (2021) 1--16.

\bibitem{israt}
I.~Z. Era, Prediction of tensile behaviors of l-ded 316 stainless steel parts using machine learning, Graduate Theses, Dissertations, and Problem Reports (2021).

\bibitem{1}
D.~Greitemeier, C.~D. Donne, F.~Syassen, J.~Eufinger, T.~Melz, Effect of surface roughness on fatigue performance of additive manufactured ti–6al–4v, Materials Science and Technology 32~(7) (2016) 629--634.

\bibitem{2}
X.~Tan, Y.~Kok, Y.~J. Tan, M.~Descoins, D.~Mangelinck, S.~B. Tor, K.~F. Leong, C.~K. Chua, Graded microstructure and mechanical properties of additive manufactured ti--6al--4v via electron beam melting, Acta Materialia 97 (2015) 1--16.

\bibitem{3}
S.~L. Lu, H.~Tang, Y.~Ning, N.~Liu, D.~StJohn, M.~Qian, Microstructure and mechanical properties of long ti-6al-4v rods additively manufactured by selective electron beam melting out of a deep powder bed and the effect of subsequent hot isostatic pressing, Metallurgical and Materials Transactions A 46~(9) (2015) 3824--3834.

\bibitem{4}
O.~L. Rodriguez, P.~G. Allison, W.~R. Whittington, D.~K. Francis, O.~G. Rivera, K.~Chou, X.~Gong, T.~Butler, J.~F. Burroughs, Dynamic tensile behavior of electron beam additive manufactured ti6al4v, Materials Science and Engineering: A 641 (2015) 323--327.

\bibitem{5}
A.~Mohammadhosseini, D.~Fraser, S.~Masood, M.~Jahedi, Microstructure and mechanical properties of ti--6al--4v manufactured by electron beam melting process, Materials Research Innovations 17~(sup2) (2013) s106--s112.

\bibitem{6}
N.~Hrabe, T.~Quinn, Effects of processing on microstructure and mechanical properties of a titanium alloy (ti--6al--4v) fabricated using electron beam melting (ebm), part 1: Distance from build plate and part size, Materials Science and Engineering: A 573 (2013) 264--270.

\bibitem{7}
H.~Rafi, N.~Karthik, H.~Gong, T.~L. Starr, B.~E. Stucker, Microstructures and mechanical properties of ti6al4v parts fabricated by selective laser melting and electron beam melting, Journal of materials engineering and performance 22~(12) (2013) 3872--3883.

\bibitem{8}
H.~Khalid~Rafi, N.~Karthik, T.~L. Starr, B.~E. Stucker, Mechanical property evaluation of ti-6al-4v parts made using electron beam melting, in: 2012 International Solid Freeform Fabrication Symposium, University of Texas at Austin, 2012.

\bibitem{9}
L.~Murr, E.~Esquivel, S.~Quinones, S.~Gaytan, M.~Lopez, E.~Martinez, F.~Medina, D.~Hernandez, E.~Martinez, J.~Martinez, et~al., Microstructures and mechanical properties of electron beam-rapid manufactured ti--6al--4v biomedical prototypes compared to wrought ti--6al--4v, Materials characterization 60~(2) (2009) 96--105.

\bibitem{10}
L.~Facchini, E.~Magalini, P.~Robotti, A.~Molinari, Microstructure and mechanical properties of ti-6al-4v produced by electron beam melting of pre-alloyed powders, Rapid Prototyping Journal (2009).

\bibitem{11}
M.~Koike, K.~Martinez, L.~Guo, G.~Chahine, R.~Kovacevic, T.~Okabe, Evaluation of titanium alloy fabricated using electron beam melting system for dental applications, Journal of Materials Processing Technology 211~(8) (2011) 1400--1408.

\bibitem{12}
M.~Larsson, U.~Lindhe, O.~Harrysson, Rapid manufacturing with electron beam melting (ebm)-a manufacturing revolution?, in: 2003 International Solid Freeform Fabrication Symposium, 2003.

\bibitem{13}
M.~Simonelli, Y.~Y. Tse, C.~Tuck, Effect of the build orientation on the mechanical properties and fracture modes of slm ti--6al--4v, Materials Science and Engineering: A 616 (2014) 1--11.

\bibitem{14}
S.~Leuders, M.~Th{\"o}ne, A.~Riemer, T.~Niendorf, T.~Tr{\"o}ster, H.~a. Richard, H.~Maier, On the mechanical behaviour of titanium alloy tial6v4 manufactured by selective laser melting: Fatigue resistance and crack growth performance, International Journal of Fatigue 48 (2013) 300--307.

\bibitem{15}
K.~Rekedal, D.~Liu, Fatigue life of selective laser melted and hot isostatically pressed ti-6al-4v absent of surface machining, in: 56th AIAA/ASCE/AHS/ASC Structures, Structural Dynamics, and Materials Conference, 2015, p. 0894.

\bibitem{16}
V.~Cain, L.~Thijs, J.~Van~Humbeeck, B.~Van~Hooreweder, R.~Knutsen, Crack propagation and fracture toughness of ti6al4v alloy produced by selective laser melting, Additive Manufacturing 5 (2015) 68--76.

\bibitem{17}
G.~Kasperovich, J.~Hausmann, Improvement of fatigue resistance and ductility of tial6v4 processed by selective laser melting, Journal of Materials Processing Technology 220 (2015) 202--214.

\bibitem{18}
P.~Edwards, M.~Ramulu, Fatigue performance evaluation of selective laser melted ti--6al--4v, Materials Science and Engineering: A 598 (2014) 327--337.

\bibitem{19}
S.~Leuders, T.~Lieneke, S.~Lammers, T.~Tr{\"o}ster, T.~Niendorf, On the fatigue properties of metals manufactured by selective laser melting--the role of ductility, Journal of Materials Research 29~(17) (2014) 1911--1919.

\bibitem{20}
M.~Seifi, M.~Dahar, R.~Aman, O.~Harrysson, J.~Beuth, J.~J. Lewandowski, Evaluation of orientation dependence of fracture toughness and fatigue crack propagation behavior of as-deposited arcam ebm ti-6al-4v, Jom 67~(3) (2015) 597--607.

\bibitem{21}
M.~Seifi, D.~Christiansen, J.~Beuth, O.~Harrysson, J.~J. Lewandowski, et~al., Process mapping, fracture and fatigue behavior of ti-6al-4v produced by ebm additive manufacturing, in: Proceedings of the 13th World Conference on Titanium, Vol. 232, Citeseer, 2016, pp. 1373--1377.

\bibitem{22}
M.~Seifi, I.~Ghamarian, P.~Samimi, U.~Ackelid, P.~Collins, J.~Lewandowski, Microstructure and mechanical properties of ti-48al-2cr-2nb manufactured via electron beam melting, in: Proceedings of the 13th World Conference on Titanium, John Wiley \& Sons, Inc. Hoboken, NJ, USA, 2016, pp. 1317--1322.

\bibitem{23}
H.~Rafi, N.~Karthik, H.~Gong, T.~L. Starr, B.~E. Stucker, Microstructures and mechanical properties of ti6al4v parts fabricated by selective laser melting and electron beam melting, Journal of materials engineering and performance 22~(12) (2013) 3872--3883.

\bibitem{24}
L.~Facchini, E.~Magalini, P.~Robotti, A.~Molinari, S.~H{\"o}ges, K.~Wissenbach, Ductility of a ti-6al-4v alloy produced by selective laser melting of prealloyed powders, Rapid Prototyping Journal (2010).

\bibitem{25}
H.~Gong, K.~Rafi, H.~Gu, G.~J. Ram, T.~Starr, B.~Stucker, Influence of defects on mechanical properties of ti--6al--4 v components produced by selective laser melting and electron beam melting, Materials \& Design 86 (2015) 545--554.

\bibitem{26}
T.~McLouth, Y.-W. Chang, J.~Wooten, J.-M. Yang, The effects of electron beam melting on the microstructure and mechanical properties of ti-6al-4v and gamma-tial, Microscopy and Microanalysis 21~(S3) (2015) 1177--1178.

\bibitem{27}
M.~Larsson, U.~Lindhe, O.~Harrysson, Rapid manufacturing with electron beam melting (ebm)-a manufacturing revolution?, in: 2003 International Solid Freeform Fabrication Symposium, 2003.

\bibitem{28}
T.~Vilaro, C.~Colin, J.-D. Bartout, As-fabricated and heat-treated microstructures of the ti-6al-4v alloy processed by selective laser melting, Metallurgical and materials transactions A 42~(10) (2011) 3190--3199.

\bibitem{29}
P.~Edwards, M.~Ramulu, Effect of build direction on the fracture toughness and fatigue crack growth in selective laser melted ti-6al-4 v, Fatigue \& Fracture of Engineering Materials \& Structures 38~(10) (2015) 1228--1236.

\bibitem{30}
B.~Van~Hooreweder, D.~Moens, R.~Boonen, J.-P. Kruth, P.~Sas, Analysis of fracture toughness and crack propagation of ti6al4v produced by selective laser melting, Advanced Engineering Materials 14~(1-2) (2012) 92--97.

\bibitem{31}
T.~H. Becker, C.~Scheffer, M.~Beck, Microstructure and mechanical properties of direct metal laser sintered ti-6al-4v d article, South African Journal of Industrial Engineering 26~(1) (2015) 1--10.

\bibitem{32}
B.~Vandenbroucke, J.-P. Kruth, Selective laser melting of biocompatible metals for rapid manufacturing of medical parts, Rapid Prototyping Journal (2007).

\bibitem{33}
A.~Mertens, S.~Reginster, H.~Paydas, Q.~Contrepois, T.~Dormal, O.~Lemaire, J.~Lecomte-Beckers, Mechanical properties of alloy ti--6al--4v and of stainless steel 316l processed by selective laser melting: influence of out-of-equilibrium microstructures, Powder Metallurgy 57~(3) (2014) 184--189.

\bibitem{34}
D.~A. Hollander, M.~Von~Walter, T.~Wirtz, R.~Sellei, B.~Schmidt-Rohlfing, O.~Paar, H.-J. Erli, Structural, mechanical and in vitro characterization of individually structured ti--6al--4v produced by direct laser forming, Biomaterials 27~(7) (2006) 955--963.

\bibitem{35}
C.~Qiu, N.~J. Adkins, M.~M. Attallah, Microstructure and tensile properties of selectively laser-melted and of hiped laser-melted ti--6al--4v, Materials Science and Engineering: A 578 (2013) 230--239.

\bibitem{36}
P.~Kobryn, S.~Semiatin, Mechanical properties of laser-deposited ti-6al-4v, in: 2001 International Solid Freeform Fabrication Symposium, 2001.

\bibitem{37}
J.~Yu, M.~Rombouts, G.~Maes, F.~Motmans, Material properties of ti6al4 v parts produced by laser metal deposition, Physics Procedia 39 (2012) 416--424.

\bibitem{38}
S.~Zhang, X.~Lin, J.~Chen, W.~Huang, Heat-treated microstructure and mechanical properties of laser solid forming ti-6al-4v alloy, Rare metals 28~(6) (2009) 537--544.

\bibitem{39}
J.~Alcisto, A.~Enriquez, H.~Garcia, S.~Hinkson, T.~Steelman, E.~Silverman, P.~Valdovino, H.~Gigerenzer, J.~Foyos, J.~Ogren, et~al., Tensile properties and microstructures of laser-formed ti-6al-4v, Journal of materials engineering and performance 20~(2) (2011) 203--212.

\bibitem{40}
G.~Dinda, L.~Song, J.~Mazumder, Fabrication of ti-6al-4v scaffolds by direct metal deposition, Metallurgical and Materials Transactions A 39~(12) (2008) 2914--2922.

\bibitem{41}
B.~E. Carroll, T.~A. Palmer, A.~M. Beese, Anisotropic tensile behavior of ti--6al--4v components fabricated with directed energy deposition additive manufacturing, Acta Materialia 87 (2015) 309--320.

\bibitem{42}
Y.~Zhai, H.~Galarraga, D.~A. Lados, Microstructure evolution, tensile properties, and fatigue damage mechanisms in ti-6al-4v alloys fabricated by two additive manufacturing techniques, Procedia Engineering 114 (2015) 658--666.

\bibitem{43}
F.~G. Arcella, F.~Froes, Producing titanium aerospace components from powder using laser forming, Jom 52~(5) (2000) 28--30.

\bibitem{44}
G.~K. Lewis, E.~Schlienger, Practical considerations and capabilities for laser assisted direct metal deposition, Materials \& Design 21~(4) (2000) 417--423.

\bibitem{45}
M.~L. Griffith, M.~T. Ensz, J.~D. Puskar, C.~V. Robino, J.~A. Brooks, J.~A. Philliber, J.~E. Smugeresky, W.~Hofmeister, Understanding the microstructure and properties of components fabricated by laser engineered net shaping (lens), MRS Online Proceedings Library (OPL) 625 (2000).

\bibitem{46}
A.~J. Sterling, B.~Torries, N.~Shamsaei, S.~M. Thompson, D.~W. Seely, Fatigue behavior and failure mechanisms of direct laser deposited ti--6al--4v, Materials Science and Engineering: A 655 (2016) 100--112.

\bibitem{47}
L.~L{\"o}ber, F.~P. Schimansky, U.~K{\"u}hn, F.~Pyczak, J.~Eckert, Selective laser melting of a beta-solidifying tnm-b1 titanium aluminide alloy, Journal of Materials Processing Technology 214~(9) (2014) 1852--1860.

\bibitem{48}
L.~L{\"o}ber, S.~Biamino, U.~Ackelid, S.~Sabbadini, P.~Epicoco, P.~Fino, J.~Eckert, Comparison off selective laser and electron beam melted titanium aluminides, in: 2011 International Solid Freeform Fabrication Symposium, University of Texas at Austin, 2011.

\bibitem{49}
A.~B. Spierings, T.~L. Starr, K.~Wegener, Fatigue performance of additive manufactured metallic parts, Rapid prototyping journal (2013).

\bibitem{50}
E.~Jelis, M.~Clemente, S.~Kerwien, N.~M. Ravindra, M.~R. Hespos, Metallurgical and mechanical evaluation of 4340 steel produced by direct metal laser sintering, Jom 67~(3) (2015) 582--589.

\bibitem{51}
I.~Tolosa, F.~Garciand{\'\i}a, F.~Zubiri, F.~Zapirain, A.~Esnaola, Study of mechanical properties of aisi 316 stainless steel processed by “selective laser melting”, following different manufacturing strategies, The International Journal of Advanced Manufacturing Technology 51~(5) (2010) 639--647.

\bibitem{52}
L.~E. Murr, E.~Martinez, S.~Gaytan, D.~Ramirez, B.~Machado, P.~Shindo, J.~Martinez, F.~Medina, J.~Wooten, D.~Ciscel, et~al., Microstructural architecture, microstructures, and mechanical properties for a nickel-base superalloy fabricated by electron beam melting, Metallurgical and Materials Transactions A 42~(11) (2011) 3491--3508.

\bibitem{53}
O.~Scott-Emuakpor, J.~Schwartz, T.~George, C.~Holycross, C.~Cross, J.~Slater, Bending fatigue life characterisation of direct metal laser sintering nickel alloy 718, Fatigue \& Fracture of Engineering Materials \& Structures 38~(9) (2015) 1105--1117.

\bibitem{54}
J.~Str{\"o}{\ss}ner, M.~Terock, U.~Glatzel, Mechanical and microstructural investigation of nickel-based superalloy in718 manufactured by selective laser melting (slm), Advanced Engineering Materials 17~(8) (2015) 1099--1105.

\bibitem{55}
K.~Kempen, L.~Thijs, J.~Van~Humbeeck, J.-P. Kruth, Processing alsi10mg by selective laser melting: parameter optimisation and material characterisation, Materials Science and Technology 31~(8) (2015) 917--923.

\bibitem{56}
S.~Siddique, M.~Imran, E.~Wycisk, C.~Emmelmann, F.~Walther, Influence of process-induced microstructure and imperfections on mechanical properties of alsi12 processed by selective laser melting, Journal of Materials Processing Technology 221 (2015) 205--213.

\bibitem{57}
T.~B. Sercombe, X.~Li, Selective laser melting of aluminium and aluminium metal matrix composites, Materials Technology 31~(2) (2016) 77--85.

\bibitem{58}
I.~Rosenthal, A.~Stern, N.~Frage, Microstructure and mechanical properties of alsi10mg parts produced by the laser beam additive manufacturing (am) technology, Metallography, Microstructure, and Analysis 3~(6) (2014) 448--453.

\bibitem{59}
X.~Wang, L.~Zhang, M.~Fang, T.~B. Sercombe, The effect of atmosphere on the structure and properties of a selective laser melted al--12si alloy, Materials Science and Engineering: A 597 (2014) 370--375.

\bibitem{60}
X.~Li, X.~Wang, M.~Saunders, A.~Suvorova, L.~Zhang, Y.~Liu, M.~Fang, Z.~Huang, T.~B. Sercombe, A selective laser melting and solution heat treatment refined al--12si alloy with a controllable ultrafine eutectic microstructure and 25\% tensile ductility, Acta Materialia 95 (2015) 74--82.

\bibitem{61}
N.~Read, W.~Wang, K.~Essa, M.~M. Attallah, Selective laser melting of alsi10mg alloy: Process optimisation and mechanical properties development, Materials \& Design (1980-2015) 65 (2015) 417--424.

\bibitem{62}
D.~Manfredi, F.~Calignano, M.~Krishnan, R.~Canali, E.~P. Ambrosio, E.~Atzeni, From powders to dense metal parts: Characterization of a commercial alsimg alloy processed through direct metal laser sintering, Materials 6~(3) (2013) 856--869.

\bibitem{63}
C.~Song, Y.~Yang, Y.~Wang, D.~Wang, J.~Yu, Research on rapid manufacturing of cocrmo alloy femoral component based on selective laser melting, The International Journal of Advanced Manufacturing Technology 75~(1) (2014) 445--453.

\bibitem{64}
R.~Kircher, A.~Christensen, K.~Wurth, Electron beam melted (ebm) co-cr-mo alloy for orthopaedic implant applications, in: 2009 International Solid Freeform Fabrication Symposium, University of Texas at Austin, 2009.

\bibitem{65}
T.~Tarasova, A.~Nazarov, M.~Prokof’ev, Effect of the regimes of selective laser melting on the structure and physicomechanical properties of cobalt-base superalloys, The Physics of Metals and Metallography 116~(6) (2015) 601--605.

\bibitem{66}
K.~Wei, M.~Gao, Z.~Wang, X.~Zeng, Effect of energy input on formability, microstructure and mechanical properties of selective laser melted az91d magnesium alloy, Materials Science and Engineering: A 611 (2014) 212--222.

\bibitem{67}
Y.~Ma, D.~Cuiuri, N.~Hoye, H.~Li, Z.~Pan, The effect of location on the microstructure and mechanical properties of titanium aluminides produced by additive layer manufacturing using in-situ alloying and gas tungsten arc welding, Materials Science and Engineering: A 631 (2015) 230--240.

\bibitem{68}
G.~Sun, R.~Zhou, J.~Lu, J.~Mazumder, Evaluation of defect density, microstructure, residual stress, elastic modulus, hardness and strength of laser-deposited aisi 4340 steel, Acta Materialia 84 (2015) 172--189.

\bibitem{69}
A.~Yadollahi, N.~Shamsaei, S.~M. Thompson, D.~W. Seely, Effects of process time interval and heat treatment on the mechanical and microstructural properties of direct laser deposited 316l stainless steel, Materials Science and Engineering: A 644 (2015) 171--183.

\bibitem{70}
B.~Baufeld, Mechanical properties of inconel 718 parts manufactured by shaped metal deposition (smd), Journal of materials engineering and performance 21~(7) (2012) 1416--1421.

\bibitem{71}
P.~Blackwell, The mechanical and microstructural characteristics of laser-deposited in718, Journal of materials processing technology 170~(1-2) (2005) 240--246.

\bibitem{72}
X.~Zhao, J.~Chen, X.~Lin, W.~Huang, Study on microstructure and mechanical properties of laser rapid forming inconel 718, Materials Science and Engineering: A 478~(1-2) (2008) 119--124.

\bibitem{73}
R.~K. Bird, J.~Hibberd, Tensile properties and microstructure of inconel 718 fabricated with electron beam freeform fabrication (ebf (sup 3)), Tech. rep. (2009).

\bibitem{74}
W.~A. Tayon, R.~N. Shenoy, M.~R. Redding, R.~Keith~Bird, R.~A. Hafley, Correlation between microstructure and mechanical properties in an inconel 718 deposit produced via electron beam freeform fabrication, Journal of Manufacturing Science and Engineering 136~(6) (2014).

\bibitem{75}
Y.-N. Zhang, X.~Cao, P.~Wanjara, M.~Medraj, Tensile properties of laser additive manufactured inconel 718 using filler wire, Journal of Materials Research 29~(17) (2014) 2006--2020.

\bibitem{76}
H.~Qi, M.~Azer, A.~Ritter, Studies of standard heat treatment effects on microstructure and mechanical properties of laser net shape manufactured inconel 718, Metallurgical and Materials Transactions A 40~(10) (2009) 2410--2422.

\bibitem{77}
X.~Cao, B.~Rivaux, M.~Jahazi, J.~Cuddy, A.~Birur, Effect of pre-and post-weld heat treatment on metallurgical and tensile properties of inconel 718 alloy butt joints welded using 4 kw nd: Yag laser, Journal of Materials Science 44~(17) (2009) 4557--4571.

\bibitem{78}
J.~Gu, B.~Cong, J.~Ding, S.~W. Williams, Y.~Zhai, Wire+ arc additive manufacturing of aluminum, in: 2014 International Solid Freeform Fabrication Symposium, University of Texas at Austin, 2014.

\bibitem{79}
E.~Kim, Y.-J. Shin, S.-H. Ahn, The effects of moisture and temperature on the mechanical properties of additive manufacturing components: fused deposition modeling, Rapid Prototyping Journal (2016).

\bibitem{80}
B.~Baufeld, O.~Van~der Biest, R.~Gault, Additive manufacturing of ti--6al--4v components by shaped metal deposition: microstructure and mechanical properties, Materials \& Design 31 (2010) S106--S111.

\bibitem{81}
A.~Popovich, V.~Sufiiarov, I.~Polozov, E.~Borisov, D.~Masaylo, A.~Orlov, Microstructure and mechanical properties of additive manufactured copper alloy, Materials Letters 179 (2016) 38--41.

\bibitem{82}
J.~Chac{\'o}n, M.~A. Caminero, E.~Garc{\'\i}a-Plaza, P.~J. N{\'u}nez, Additive manufacturing of pla structures using fused deposition modelling: Effect of process parameters on mechanical properties and their optimal selection, Materials \& Design 124 (2017) 143--157.

\bibitem{83}
Y.~Kok, X.~P. Tan, P.~Wang, M.~Nai, N.~H. Loh, E.~Liu, S.~B. Tor, Anisotropy and heterogeneity of microstructure and mechanical properties in metal additive manufacturing: A critical review, Materials \& Design 139 (2018) 565--586.

\bibitem{84}
H.~Fayazfar, M.~Salarian, A.~Rogalsky, D.~Sarker, P.~Russo, V.~Paserin, E.~Toyserkani, A critical review of powder-based additive manufacturing of ferrous alloys: Process parameters, microstructure and mechanical properties, Materials \& Design 144 (2018) 98--128.

\bibitem{87}
A.~Moshki, M.~R. Hajighasemi, A.~A. Atai, E.~Jebellat, A.~Ghazavizadeh, Optimal design of 3d architected porous/nonporous microstructures of multifunctional multiphase composites for maximized thermomechanical properties, Comput Mech 69 (2022) 979–996.

\bibitem{86}
K.~Schmidtke, F.~Palm, A.~Hawkins, C.~Emmelmann, Process and mechanical properties: applicability of a scandium modified al-alloy for laser additive manufacturing, Physics Procedia 12 (2011) 369--374.

\bibitem{85}
H.~Hack, R.~Link, E.~Knudsen, B.~Baker, S.~Olig, Mechanical properties of additive manufactured nickel alloy 625, Additive Manufacturing 14 (2017) 105--115.

\bibitem{plotdigitizer}
Plotdigitizer: Version 2.6.8 (2015).

\bibitem{10.1115/1.4035420}
R.~E. Laureijs, J.~B. Roca, S.~P. Narra, C.~Montgomery, J.~L. Beuth, E.~R.~H. Fuchs, {Metal Additive Manufacturing: Cost Competitive Beyond Low Volumes}, Journal of Manufacturing Science and Engineering 139~(8), 081010 (05 2017).

\bibitem{mendeleev}
{mendeleev}, \url{https://mendeleev.readthedocs.io/en/stable/} (2018).

\bibitem{CARRUTHERS1988351}
C.~Carruthers, H.~Teitelbaum, The linear mixture rule in chemical kinetics. ii. thermal dissociation of diatomic molecules, Chemical Physics 127~(1) (1988) 351--362.

\bibitem{RF}
L.~Breiman, Random forests., Machine Learning 45 (2001) 5–32.

\bibitem{GB}
J.~Friedman, Greedy function approximation: A gradient boosting machine., Annals of Statistics 29 (2001) 1189--1232.

\bibitem{svm}
C.~Cortes, V.~Vapnik, Support-vector networks., Machine Learning 20 (1995) 273--297.

\bibitem{Rasmussen2004}
C.~E. Rasmussen, Gaussian processes in machine learning, in: Summer school on machine learning, Springer, 2003, pp. 63--71.

\bibitem{lasso}
R.~Tibshirani, Regression shrinkage and selection via the lasso, Journal of the Royal Statistical Society: Series B (Methodological) 58~(1) (1996) 267--288.

\bibitem{Ridge}
A.~E. Hoerl, R.~W. Kennard, Ridge regression: Biased estimation for nonorthogonal problems, Technometrics 12~(1) (1970) 55--67.

\bibitem{NN}
G.~E. Hinton, S.~Osindero, Y.-W. Teh, {A Fast Learning Algorithm for Deep Belief Nets}, Neural Computation 18~(7) (2006) 1527--1554.

\bibitem{Chen2016}
T.~Chen, C.~Guestrin, “xgboost: A scalable tree boosting system.”, Proceedings of the 22nd ACM SIGKDD International Conference on Knowledge Discovery and Data Mining (2016) 785–94.

\bibitem{JMLR:v13:bergstra12a}
J.~Bergstra, Y.~Bengio, Random search for hyper-parameter optimization, Journal of Machine Learning Research 13~(10) (2012) 281--305.

\bibitem{bergstra2011algorithms}
J.~Bergstra, R.~Bardenet, Y.~Bengio, B.~K{\'e}gl, Algorithms for hyper-parameter optimization, Advances in neural information processing systems 24 (2011).

\bibitem{hyperopt}
J.~Bergstra, D.~Yamins, D.~D. Cox, Making a science of model search: Hyperparameter optimization in hundreds of dimensions for vision architectures, in: Proceedings of the 30th International Conference on International Conference on Machine Learning - Volume 28, ICML'13, JMLR.org, 2013, p. I–115–I–123.

\bibitem{probst2019hyperparameters}
P.~Probst, M.~N. Wright, A.-L. Boulesteix, Hyperparameters and tuning strategies for random forest, Wiley Interdisciplinary Reviews: Data Mining and Knowledge Discovery 9~(3) (2019) e1301.

\bibitem{scholkopf2002learning}
B.~Sch{\"o}lkopf, A.~J. Smola, F.~Bach, et~al., Learning with kernels: support vector machines, regularization, optimization, and beyond, MIT press, 2002.

\bibitem{DUAN200341}
K.~Duan, S.~Keerthi, A.~N. Poo, Evaluation of simple performance measures for tuning svm hyperparameters, Neurocomputing 51 (2003) 41--59.

\bibitem{liu2020gaussian}
H.~Liu, Y.-S. Ong, X.~Shen, J.~Cai, When gaussian process meets big data: A review of scalable gps, IEEE transactions on neural networks and learning systems 31~(11) (2020) 4405--4423.

\bibitem{wei2014evading}
Y.~Wei, Y.~Li, L.~Zhu, Y.~Liu, X.~Lei, G.~Wang, Y.~Wu, Z.~Mi, J.~Liu, H.~Wang, et~al., Evading the strength--ductility trade-off dilemma in steel through gradient hierarchical nanotwins, Nature communications 5~(1) (2014) 1--8.

\bibitem{HRABE2013271}
N.~Hrabe, T.~Quinn, Effects of processing on microstructure and mechanical properties of a titanium alloy (ti–6al–4v) fabricated using electron beam melting (ebm), part 2: Energy input, orientation, and location, Materials Science and Engineering: A 573 (2013) 271--277.

\bibitem{NIPS2017_7062}
S.~M. Lundberg, S.-I. Lee, A unified approach to interpreting model predictions, in: I.~Guyon, U.~V. Luxburg, S.~Bengio, H.~Wallach, R.~Fergus, S.~Vishwanathan, R.~Garnett (Eds.), Advances in Neural Information Processing Systems 30, Curran Associates, Inc., 2017, pp. 4765--4774.

\bibitem{lundberg2020local2global}
S.~M. Lundberg, G.~Erion, H.~Chen, A.~DeGrave, J.~M. Prutkin, B.~Nair, R.~Katz, J.~Himmelfarb, N.~Bansal, S.-I. Lee, From local explanations to global understanding with explainable ai for trees, Nature Machine Intelligence 2~(1) (2020) 2522--5839.

\bibitem{ai-feynman}
S.-M. Udrescu, M.~Tegmark, Ai feynman: A physics-inspired method for symbolic regression, Science Advances 6~(16) (2020) eaay2631.

\bibitem{Sindy}
S.~L. Brunton, J.~L. Proctor, J.~N. Kutz, Discovering governing equations from data by sparse identification of nonlinear dynamical systems, Proceedings of the National Academy of Sciences 113~(15) (2016) 3932--3937.

\bibitem{2020SciPy-NMeth}
P.~Virtanen, R.~Gommers, T.~E. Oliphant, M.~Haberland, T.~Reddy, D.~Cournapeau, E.~Burovski, P.~Peterson, W.~Weckesser, J.~Bright, S.~J. {van der Walt}, M.~Brett, J.~Wilson, K.~J. Millman, N.~Mayorov, A.~R.~J. Nelson, E.~Jones, R.~Kern, E.~Larson, C.~J. Carey, {\.I}.~Polat, Y.~Feng, E.~W. Moore, J.~{VanderPlas}, D.~Laxalde, J.~Perktold, R.~Cimrman, I.~Henriksen, E.~A. Quintero, C.~R. Harris, A.~M. Archibald, A.~H. Ribeiro, F.~Pedregosa, P.~{van Mulbregt}, {SciPy 1.0 Contributors}, {{SciPy} 1.0: Fundamental Algorithms for Scientific Computing in Python}, Nature Methods 17 (2020) 261--272.

\bibitem{trust-region}
A.~R. Conn, N.~I.~M. Gould, P.~L. Toint, Trust Region Methods, Society for Industrial and Applied Mathematics, 2000.

\end{thebibliography}

\end{document}